\documentclass[twoside,11pt,letterpaper]{article}

\usepackage{jmlr2e}

\usepackage[latin1]{inputenc}
\usepackage[T1]{fontenc}
\usepackage{amsmath,amsfonts}
\usepackage[footnotesize,raggedright,hang,tight]{subfigure}
\usepackage{url}

\DeclareMathOperator{\Kff}{\mathbf{K}_{f,f}}
\DeclareMathOperator{\iKff}{ \mathbf{K}_{f,f}^{-1}}

\DeclareMathOperator{\x}{\mathbf{x}}
\DeclareMathOperator{\X}{\mathbf{X}}
\DeclareMathOperator{\f}{\mathbf{f}}
\DeclareMathOperator{\y}{\mathbf{y}}
\DeclareMathOperator{\D}{\mathcal{D}}
\DeclareMathOperator{\Sq}{\boldsymbol{\Sigma}}
\DeclareMathOperator{\mq}{\boldsymbol{\mu}}

\DeclareMathOperator{\St}{\tilde{\boldsymbol{\Sigma}}}
\DeclareMathOperator{\mt}{\tilde{\boldsymbol{\mu}}}

\DeclareMathOperator{\nut}{\tilde{ \boldsymbol{\nu} }}

\DeclareMathOperator{\lamb}{ \boldsymbol{\lambda} }

\DeclareMathOperator{\Tr}{^{\text{T}}}
\DeclareMathOperator{\I}{\mathbf{I}}

\DeclareMathOperator{\GP}{\mathcal{GP}}
\DeclareMathOperator{\N}{\mathcal{N}}

\DeclareMathOperator{\E}{E}

\DeclareMathOperator{\KL}{KL}

\DeclareMathOperator*{\argmax}{arg\,max}
\DeclareMathOperator*{\argmin}{arg\,min}
\newcommand{\mb}[1]{\mathbf{#1}}

\newcommand{\Tra}[1]{ #1^\text{T} }

\ShortHeadings{Gaussian Process Regression with a Student-$t$ Likelihood}{Pasi Jylänki, Jarno Vanhatalo and Aki Vehtari}
\firstpageno{1}

\begin{document}
\thispagestyle{empty}

\title{Robust Gaussian Process Regression with a Student-$t$ Likelihood}

\author{\name Pasi Jylänki \email pasi.jylanki@aalto.fi \\
       \addr Department of Biomedical Engineering and Computational Science\\
       Aalto University\\
       P.O. Box 12200, FI-00076 AALTO, Espoo, Finland
       \AND
       \name Jarno Vanhatalo \email jarno.vanhatalo@helsinki.fi \\
       \addr Department of Environmental Sciences\\
       University of Helsinki\\
       P.O. Box 65, FI-00014 University of Helsinki, Finland
       \AND
       \name Aki Vehtari \email aki.vehtari@aalto.fi \\
       \addr Department of Biomedical Engineering and Computational Science\\
       Aalto University\\
       P.O. Box 12200, FI-00076 AALTO, Espoo, Finland}

\editor{}

\maketitle

\begin{abstract}
  This paper considers the robust and efficient implementation of
  Gaussian process regression with a Student-$t$ observation model.
  The challenge with the Student-$t$ model is the analytically
  intractable inference which is why several approximative methods
  have been proposed. The expectation propagation (EP) has been
  found to be a very accurate method in many empirical studies but
  the convergence of the EP is known to be problematic with models
  containing non-log-concave site functions such as the Student-$t$
  distribution. In this paper we illustrate the situations where
  the standard EP fails to converge and review different
  modifications and alternative algorithms for improving the
  convergence. We demonstrate that convergence problems may occur
  during the type-II maximum a posteriori (MAP) estimation of the
  hyperparameters and show that the standard EP may not converge in
  the MAP values in some difficult cases. We present a robust
  implementation which relies primarily on parallel EP updates and
  utilizes a moment-matching-based double-loop algorithm with
  adaptively selected step size in difficult cases. The predictive
  performance of the EP is compared to the Laplace, variational
  Bayes, and Markov chain Monte Carlo approximations.
%
%
%
%
\end{abstract}


\section{Introduction}





In many regression problems observations may include outliers which deviate
strongly from the other members of the sample. Such outliers may occur, for
example, because of failures in the measurement process or absence of certain
relevant explanatory variables in the model. In such cases, a robust observation
model is required.
%

Robust inference has been studied extensively. \citet{DeFinetti:1961} described how
Bayesian inference on the mean of a random sample, assuming a suitable observation
model, naturally leads to giving less weight to outlying observations. However, in
contrast to simple rejection of outliers, the posterior depends on all data but in
the limit, as the separation between the outliers and the rest of the data
increases, the effect of outliers becomes negligible. More theoretical results on
this kind of outlier rejection were presented by \citet{Dawid:1973} who gave
sufficient conditions on the observation model $p(y|\theta)$ and the prior
distribution $p(\theta)$ of an unknown location parameter $\theta$, which ensure
that the posterior expectation of a given function $m(\theta)$ tends to the prior
as $y \rightarrow \infty$. He also stated that the Student-$t$ distribution
combined with a normal prior has this property.

A more formal definition of robustness was given by
\citet{OHagan:1979} in terms of an outlier-prone observation model.
The observation model is defined to be outlier-prone of order $n$, if
$p(\theta|y_1,...,y_{n+1}) \rightarrow p(\theta| y_1,...,y_{n})$ as
$y_{n+1} \rightarrow \infty$. 
That is, the effect of a single conflicting observation to the posterior becomes
asymptotically negligible as the observation approaches infinity.
\citet{OHagan:1979} showed that the Student-$t$ distribution is outlier prone of
order 1, and that it can reject up to $m$ outliers if there are at least $2m$
observations altogether. This contrasts heavily with the commonly used Gaussian
observation model in which each observation influences the posterior no matter how
far it is from the others. In the nonlinear Gaussian process (GP) regression
context the outlier rejection is more complicated and one may consider the
posterior distribution of the unknown function values $f_i=f(x_i)$ locally near
some input locations $x_i$. Depending on the smoothness properties defined through
the prior on $f_i$, $m$ observations can be rejected locally if there are at least
$2m$ data points nearby. However, already two conflicting data points can render
the posterior distribution multimodal making the posterior inference challenging
(these issues will be illustrated in the upcoming sections).


In this work, we adopt the Student-$t$ observation model for GP regression because
of its good robustness properties which can be altered continuously from a very
heavy tailed distribution to the Gaussian model with the degrees of freedom
parameter. This allows the extent of robustness to be determined from the data
through hyperparameter inference. The Student-$t$ observation model was studied in
linear regression by \citet{West:1984} and \citet{Geweke:1993}, and
\citet{Neal:1997} introduced it for GP regression. Other robust observation models
which have been utilized in the GP regression include, for example, mixtures of
Gaussians \citep{Kuss:2006, Stegle+Fallert+MacKay+Brage:2008}, the Laplace
distribution \citep{Kuss:2006}, and input dependent observation models
\citep{Goldberg+Williams+Bishop:1998, Naish-Guzman+Holden:2008b}.

The challenge with the Student-$t$ model is the analytically intractable inference.
A common approach has been to use the scale-mixture representation of the
Student-$t$ distribution \citep{Geweke:1993}, which enables Gibbs sampling
\citep{Geweke:1993, Neal:1997}, and a factorizing variational approximation (fVB)
for the posterior inference \citep{Tipping+Lawrence:2005,Kuss:2006}. Recently
\citet{Vanhatalo+Jylanki+Vehtari:2009} compared the fVB with the Laplace
approximation (see, e.g., \citet{Rasmussen+Williams:2006}) and showed that
Laplace's method provided slightly better predictive performance with less
computational burden. They also showed that the fVB tends to underestimate the
posterior uncertainties of the predictions because it assumes the scales and the
unknown function values a posteriori independent. Another variational approach
called variational bounds (VB)
%
%
is available in the GPML software package \citep{GPML}. The method is based on
forming an un-normalized Gaussian lower bound for each non-Gaussian likelihood term
independently (see \cite{Nickisch+Rasmussen:2008} for details and comparisons in GP
classification). Yet another related variational approach is described by
\citet{Opper+Archambeau:2009} who studied the Cauchy observation model (Student-$t$
with degrees of freedom 1). This method is similar to the KL-divergence
minimization approach (KL) described by \citet{Nickisch+Rasmussen:2008} and the VB
approach can be regarded as a special case of KL. The extensive comparisons by
\citet{Nickisch+Rasmussen:2008} in GP classification suggest that the VB provides
better predictive performance than the Laplace approximation but worse marginal
likelihood estimates than the KL or the expectation propagation (EP)
\citep{Minka:2001}. According to the comparisons of
\citet{Nickisch+Rasmussen:2008}, EP is the method of choice since it is much faster
than KL, at least in GP classification. The problem with the EP, however, is that
the Student-$t$ likelihood is not log-concave which may lead to convergence
problems
\citep{Seeger:2008}. 

In this paper, we focus on establishing a robust EP implementation for the
Student-$t$ observation model. We illustrate the convergence problems of the
standard EP with simple one-dimensional regression examples and discuss how
damping, fractional EP updates (or power EP) \citep{Minka:2004,Seeger:2005}, and
double-loop algorithms \citep{Heskes:2002} can be used to improve the convergence.
We present a robust implementation which relies primarily on parallel EP updates
\citep[see e.g.,][]{Gerven:2009} and utilizes a moment-matching-based double-loop
algorithm with adaptively selected step size to find stationary solutions in
difficult cases. We show that the implementation enables a robust type-II maximum a
posteriori (MAP) estimation of the hyperparameters based on the approximative
marginal likelihood.
%
%
The proposed implementation is compared to the Laplace approximation, fVB, VB, and
Markov chain Monte Carlo (MCMC) using one simulated and three real-world data sets.
%

\section{Gaussian Process Regression with Student-$t$ Observation Model}

We will consider a regression problem, with scalar observations $y_i = f(\x_i) +
\epsilon_i, i=1,...,n$ at input locations $\mb{X}=\{\mb{x}_i\}_{i=1}^n$, and where
the observation errors $\epsilon_1,...,\epsilon_n$ are zero-mean exchangeable
random variables. The object of inference is the latent function $f(\x):\Re^d
\rightarrow \Re$, which is given a Gaussian process prior
\begin{equation}
f(\x)|\theta \sim \GP\left(m(\x),  k(\x,\x'|\theta)\right), \label{GP_prior}
\end{equation}
where $m(\x)$ and $k(\x,\x'|\theta)$ are the mean and covariance functions of the
process controlled by hyperparameters $\theta$. For notational simplicity we will
assume a zero mean GP. By definition, a Gaussian process prior implies that any
finite subset of latent variables, $\mb{f}=\{f(\mb{x}_i)\}_{i=1}^n$, has a
multivariate Gaussian distribution. In particular, at the observed input locations
$\mb{X}$ the latent variables are distributed as
$p(\f|\mb{X},\theta)=\N(\f|\mb{0},\Kff)$, where $\Kff$ is a covariance matrix with
entries $[\Kff]_{ij}=k(\x_i,\x_j|\theta)$. The covariance function encodes the
prior assumptions on the latent function, such as the smoothness and scale of the
variation, and can be chosen freely as long as the covariance matrices which it
produces are symmetric and positive semi-definite. An example of a stationary
covariance function is the squared exponential
\begin{equation} \label{cf_sexp}
k_{\mathrm{se}}(\x_i,\x_j|\theta)=\sigma_{\mathrm{se}}^2\exp\left(-
  \sum_{k=1}^d \frac{(x_{i,k}-x_{j,k})^2}{2l_k^2}\right),
\end{equation}
%
where $\theta = \{\sigma_{\mathrm{se}}^2, l_1,...,l_d\}$, $\sigma_{\mathrm{se}}^2$
is a magnitude parameter which scales the overall variation of the unknown
function, and $l_k$ is a length-scale parameter which governs how fast the
correlation decreases as the distance increases in the input dimension $k$.

The traditional assumption is that given $\f$ the error terms $\epsilon_i$ are
i.i.d. Gaussian: $\epsilon_i \sim \N(0,\sigma^2)$. In this case, the marginal
likelihood $p(\mb{y}|\mb{X}, \mb{\theta}, \sigma^2)$ and the conditional posterior
of the latent variables $p(\f|\mathcal{D},\theta, \sigma^2)$, where
$\mathcal{D}=\{\y,\mb{X}\}$, have an analytical solution. This is computationally
convenient since approximate methods are needed only for the inference on the
hyperparameters $\theta$ and $\sigma^2$. However, the limitation with the Gaussian
model is its non-robustness. The robust Student-$t$ observation model
\begin{equation*}
p(y_i|f_i,\sigma^2, \nu) =
\frac{\Gamma((\nu+1)/2)}{\Gamma(\nu/2)\sqrt{\nu\pi}\sigma}\left(1 +
  \frac{(y_i-f_i)^2}{\nu\sigma^2} \right)^{-(\nu+1)/2},
\end{equation*}
where $\nu$ is the degrees of freedom and $\sigma$ the scale parameter
\citep{Gelman+Carlin+Stern+Rubin:2004}, is computationally challenging. The
marginal likelihood and the conditional posterior $p(\f|\mathcal{D},\theta,
\sigma^2)$ are not anymore analytically tractable but require some method for
approximate inference.


\section{Approximate Inference for the Student-$t$ Model}

In this section, we review the approximate inference methods considered in this
paper. First we give a short description of the MCMC and the Laplace approximation,
as well as the two variational methods, fVB and VB. Then we give a more detailed
description of the EP algorithm and review ways to improve the convergence in more
difficult problems.


\subsection{Markov Chain Monte Carlo}

The MCMC approach is based on drawing samples from $p(\f, \theta,\sigma^2,
\nu|\mathcal{D})$ and using these samples to represent the posterior distribution
and to numerically approximate integrals over the latent variables and the
hyperparameters. Instead of implementing a Markov chain sampler directly for the
Student-$t$ model a more common approach is to use the Gibbs sampling based on the
following scale mixture representation of the Student-$t$ distribution
\vspace{-3mm}
\begin{align}
y_i | f_i & \sim \N(f_i, V_i) \nonumber\\
V_i & \sim \text{Inv-}\chi^2(\nu, \sigma^2), \label{eq_scale_mixture}
\end{align}
where each observation has its own $\text{Inv-}\chi^2$-distributed noise variance
$V_i$ \citep{Neal:1997,Gelman+Carlin+Stern+Rubin:2004}. Sampling of the
hyperparameters $\theta$ can be done with any general sampling algorithm, such as
the Slice sampling or the hybrid Monte Carlo (HMC) \citep[see,
e.g.,][]{Gelman+Carlin+Stern+Rubin:2004}. The Gibbs sampler on the scale mixture
(\ref{eq_scale_mixture}) converges often slowly and may get stuck for long times in
small values of $\sigma^2$ because of the dependence between $V_i$ and $\sigma^2$.
%
%
This can be avoided by re-parameterization $V_i = \alpha^2 U_i$, where $U_i \sim
\text{Inv-}\chi^2 (\nu,\tau^2)$, and $\log \alpha^2 \propto 1$
\citep{Gelman+Carlin+Stern+Rubin:2004}. This improves mixing of the chains and
reduces the autocorrelations but introduces an implicit prior for the scale
parameter $\sigma^2 = \alpha^2 \tau^2$ of the Student-$t$ model.

%

\subsection{Laplace Approximation}

The Laplace approximation for the conditional posterior of the latent function is
constructed from the second order Taylor expansion of $\log p(\f|\D, \theta,
\sigma^2, \nu)$ around the mode $\hat{\f}$, which gives a Gaussian approximation to
the conditional posterior
\begin{equation}
p(\f|\D,\theta,\sigma^2, \nu) \approx q(\f|\mathcal{D},\theta,\sigma^2, \nu) =
\N(\f|\hat{\f}, \Sq_{\text{LA}}),
\label{eq_LA_post}
\end{equation}
where $\hat{\f}=\argmax_{\f} p(\f|\mathcal{D},\theta,\sigma^2, \nu)$
\citep{Rasmussen+Williams:2006}. $\Sq_{\text{LA}}^{-1}$ is the Hessian of the
negative log conditional posterior at the mode, that is,
\begin{equation}
\mb{\Sigma}_{\text{LA}}^{-1} = -\nabla\nabla \log
p(\f|\mathcal{D},\theta,\sigma^2, \nu)|_{\f=\hat{\f}} = \iKff + \mb{W},
\label{Hessian}
\end{equation}
where $\mb{W}$ is a diagonal matrix with entries $\mb{W}_{ii} =
\nabla_{f_i}\nabla_{f_i} \log p(y|f_i,\sigma^2, \nu)|_{f_i=\hat{f}_i}$.

The inference in the hyperparameters is conducted by doing a Laplace approximation
to the marginal likelihood $p(\mb{y}|\mb{X},\mb{\theta},\sigma^2,\nu)$ and
searching for the maximum a posterior (MAP) estimate for the hyperparameters
\begin{equation*}
\{\hat{\theta}, \hat{\sigma}^2,\hat{\nu}\} = \argmax_{\theta,\sigma^2,\nu}
\log q(\theta,\sigma^2,\nu|\mathcal{D}) = \argmax_{\theta,\sigma^2,\nu}
\left[ \log q(\mb{y}|\mb{X},\mb{\theta},\sigma^2,\nu) + \log p(\theta,\sigma^2,\nu) \right],
\end{equation*}
where $p(\theta,\sigma^2,\nu)$ is the prior of the hyperparameters.
%
%
%
%
%
The gradients of the approximate log marginal likelihood can be solved
analytically, which enables the MAP estimation of the hyperparameters with gradient
based optimization methods. Following \citet{Williams+Barber:1998} the
approximation scheme is called the Laplace method, but essentially the same
approach is named Gaussian approximation by \citet{Rue+Martino+Chopin:2009} in
their Integrated nested Laplace approximation (INLA) software package for Gaussian
Markov random field models \citep{Vanhatalo+Jylanki+Vehtari:2009}, \citep[see
also][]{Tierney+Kadane:1986}.

The implementation of the Laplace algorithm for this particular model requires care
since the Student-$t$ likelihood is not log-concave and thus
$p(\f|\mathcal{D},\theta,\sigma^2, \nu)$ may be multimodal. The standard
implementation presented by \citet{Rasmussen+Williams:2006} requires some
modifications which are discussed in detail by
\citet{Vanhatalo+Jylanki+Vehtari:2009}. Later on Hannes Nickisch proposed a
sligthly different implementation (personal communication), which is at the moment
in use in the GPML software package \citep{GPML}.

\subsection{Factorizing variational approximation (fVB)}

The scale-mixture decomposition (\ref{eq_scale_mixture}) enables a computationally
convenient variational approximation if the latent values $\f$ and the residual
variance terms $\mathbf{V} = [V_1,...,V_n]$ are assumed a posteriori independent:
\begin{equation} \label{eq_fVB}
  q(\f,\mathbf{V})=q(\f) \prod_{i=1}^n q(V_i).
\end{equation}
This kind of factorizing variational approximation was introduced by
\citet{Tipping:2003} to form a robust observation model for linear models within
the relevance vector machine framework. For robust Gaussian process regression with
the Student-$t$ likelihood it was applied by \citet{Kuss:2006} and essentially the
same variational approach has also been used for approximate inference on linear
models with the automatic relevance determination prior \citep[see
e.g.,][]{Tipping+Lawrence:2005}. Assuming the factorizing posterior (\ref{eq_fVB})
and minimizing the KL-divergence from $q(\f,\mathbf{V})$ to the true posterior
$p(\f,\mathbf{V}|\D,\theta,\sigma^2,\nu)$ results in a Gaussian approximation for
the latent values, and inverse-$\chi^2$ (or equivalently inverse gamma)
approximations for the residual variances $V_i$. The parameters of $q(\f)$ and
$q(V_i)$ can be estimated by maximizing a variational lower bound for the marginal
likelihood $p(\y|X,\theta,\sigma^2,\nu)$ with an expectation maximization (EM)
algorithm. In the E-step of the algorithm the lower bound is maximized with respect
to $q(\f)$ and $q(V_i)$ given the current point estimate of the hyperparameters and
in the M-step a new estimate of the hyperparameters is determined with fixed
$q(\f)$ and $q(V_i)$.

The drawback with a factorizing approximation determined by minimizing the reverse
KL-divergence is that it tends to underestimate the posterior uncertainties
\citep[see e.g.,][]{Bishop:2006}. \citet{Vanhatalo+Jylanki+Vehtari:2009} compared
fVB with the previously described Laplace and MCMC approximations, and found that
the fVB provided worse predictive variance estimates compared to the Laplace
approximation. In addition, the estimation of $\nu$ based on maximizing the
variational lower bound was found less robust with the fVB.

\subsection{Variational bounds (VB)}

This variational bounding method was introduced for binary GP classification by
\citet{Gibbs:2000} and comparisons to other approximative methods for GP
classification can be found in \citep{Nickisch+Rasmussen:2008}. The method is based
on forming a Gaussian lower bound for each likelihood term independently:
\begin{equation*}\label{eq_VB_bounds}
  p(y_i| f_i) \ge \exp(-f_i^2 /(2 \gamma_i) + b_i f_i -h(\gamma_i)/2 ),
\end{equation*}
which can be used to construct a lower bound on the marginal likelihood:
$p(\y|\X,\theta,\nu,\sigma)  \ge Z_{\text{VB}}$. With fixed hyperparameters,
$\gamma_i$ and $b_i$ can be determined by maximizing $Z_{\text{VB}}$ to obtain a
Gaussian approximation for the latent values $p(\f|\D,\theta,\nu,\sigma^2)$ and an
approximation for the marginal likelihood. With the Student-$t$ likelihood only the
scale parameters $\gamma_i$ need to be optimized because the location parameter is
determined by the corresponding observations: $b_i=y_i/\gamma_i$. Similarly to the
Laplace approximation and EP, MAP-estimation of the hyperparameters can be done by
optimizing $Z_{\text{VB}}$ with gradient based methods. In our experiments we used
the implementation available in the GPML-package \citep{GPML} augmented with the
same hyperprior definitions as with the other approximative methods.


\subsection{Expectation Propagation}

The EP algorithm is a general method for approximating integrals over functions
that factor into simple terms \citep{Minka:2001}.
%
%
%
%
It approximates the conditional posterior with
\begin{equation}\label{eq_EP_post}
q(\f|\mathcal{D}, \theta, \sigma^2, \nu) = \frac{1}{Z_{\text{EP}}}p(\f|\theta)\prod_{i=1}^n
\tilde{t}_i(f_i|\tilde{Z}_i, \tilde{\mu}_i,\tilde{\sigma}_i^2) = \N(\mq,\Sq),
\end{equation}
%
%
where $Z_{\text{EP}} \approx p(\y|\X,\theta,\sigma^2, \nu)$,
$\Sq=(\iKff+\St^{-1})^{-1}$, $\mq=\Sq \St^{-1} \mt $, $\St =
\text{diag}[\tilde{\sigma}_1^{2},...,\tilde{\sigma}_n^{2}]$, and $\mt =
[\tilde{\mu}_1,...,\tilde{\mu}_n] \Tr$. In (\ref{eq_EP_post}) the likelihood terms
$p(y_i|f_i,\sigma^2, \nu)$ are approximated by un-normalized Gaussian site
functions $\tilde{t}_i(f_i|\tilde{Z}_i, \tilde{\mu}_i,\tilde{\sigma}_i^2) =
\tilde{Z}_i \N(f_i|\tilde{\mu}_i,\tilde{\sigma}_i^2)$.

The EP algorithm updates the site parameters $\tilde{Z}_i$, $\tilde{\mu}_i$ and
$\tilde{\sigma}_i^2$ and the posterior approximation ($\ref{eq_EP_post}$)
sequentially. At each iteration ($i$), first the $i$'th site is removed from the
$i$'th marginal posterior to obtain a cavity distribution
\begin{equation}\label{eq_EP_cavity}
q_{-i}(f_i) \propto q(f_i|\mathcal{D}, \theta,\sigma^2,\nu) \tilde{t}_i(f_i)^{-1}.
\end{equation}
Then the $i$'th site is replaced with the exact likelihood term to form a tilted
distribution $\hat{p}_i (f_i) = \hat{Z}_i^{-1} q_{-i}(f_i)p(y_i|f_i)$ which is a
more refined non-Gaussian approximation to the true $i$'th marginal distribution.
Next the algorithm attempts to match the approximative posterior marginal $q(f_i)$
with $\hat{p}_i(f_i)$ by finding first a Gaussian $\hat{q}_i(f_i)$ satisfying
\begin{equation}
  \hat{q}_i(f_i) = \N(f_i| \hat{\mu}_i, \hat{\sigma}_i^2) = \argmin_{q_i}
  \KL\left(\hat{p}_i(f_i)||q_i(f_i)\right),
\end{equation}
%
%
%
%
which is equivalent to matching $\hat{\mu}_i$ and $\hat{\sigma}_i^2$ with the mean
and variance of $\hat{p}_i(f_i)$. Then the parameters of the local approximation
$\tilde{t}_i$ are updated so that the moments of $q(f_i)$ match with
$\hat{q}(f_i)$:
\begin{equation} \label{eq_moment_matching}
  q(f_i|\mathcal{D}, \theta, \sigma^2, \nu) \propto q_{-i}(f_i)\tilde{t}_i(f_i) \equiv
  \hat{Z}_i\N(f_i| \hat{\mu}_i, \hat{\sigma}_i^2).
\end{equation}
Finally, the parameters $\mq$ and $\Sq$ of the approximate posterior
\eqref{eq_EP_post} are updated according to the changes in site $\tilde{t}_i$.
These steps are repeated for all the sites at some order until convergence.
%
Since only the means and variances are needed in the Gaussian moment matching only
$\tilde{\mu}_i$ and $\tilde{\sigma}_i^2$ need to be updated during the iterations.
The normalization terms $\tilde{Z}_i$ are required for the marginal likelihood
approximation $Z_{\text{EP}} \approx p(\y|\X,\theta,\sigma^2, \nu)$ which is
computed after converge of the algorithm, and they can be determined by integrating
over $f_i$ in equation (\ref{eq_moment_matching}) which gives $\tilde{Z}_i =
\hat{Z}_i ( \int q_{-i}(f_i) \N(f_i|\tilde{\mu}_i, \tilde{\sigma}_i^2) df_i
)^{-1}$.
%
%

In the traditional EP algorithm (from now on referred to as the sequential EP), the
posterior approximation (\ref{eq_EP_post}) is updated sequentially after each
moment matching $(\ref{eq_moment_matching})$. Recently an alternative parallel
update scheme has been utilized especially in models with a very large number of
unknowns (see e.g., \citet{Gerven:2009}). In the parallel EP the site updates are
calculated with fixed posterior marginals $\mq$ and $\text{diag}(\Sq)$ for all
$\tilde{t}_i$, $i=1,...,n$, in parallel, and the posterior approximation is
refreshed only after all the sites have been updated. Although the theoretical cost
for one sweep over the sites is the same ($\mathcal{O}(n^3)$) for both the
sequential and the parallel EP, in practice one re-computation of $\Sq$ using
Cholesky decomposition is much more efficient than $n$ sequential rank-one updates.
In our experiments, the number of sweeps required for convergence was roughly the
same for both schemes in easier cases where the standard EP converges.

The marginal likelihood approximation is given by
\begin{align} \label{EP_marg_likelih_0}
\log Z_{\text{EP}} =&  -\frac{1}{2} \log|\Kff+\St|
-\frac{1}{2}\mt \Tr \left(\Kff+\St
\right)^{-1} \mt + \sum_{i=1}^n \log \hat{Z}_i(\sigma^2, \nu) +C_{\text{EP}},
\end{align}
where $C_{\text{EP}}= -\frac{n}{2} \log(2 \pi) -\sum_i \log \int q_{-i}(f_i)
\N(f_i|\tilde{\mu}_i, \tilde{\sigma}_i^2) df_i$ collects terms that are not
explicit functions of $\theta$, $\sigma^2$ or $\nu$. If the algorithm has
converged, that is, $\hat{p}_i(f_i)$ is consistent (has the same means and
variances) with $q(f_i)$ for all sites, $C_{\text{EP}}$, $\St$ and $\mt$ can be
considered constants when differentiating \eqref{EP_marg_likelih_0} with respect to
the hyperparameters \citep{Seeger:2005,Opper:2005}. This enables efficient MAP
estimation with gradient based optimization methods.

There is no guarantee of convergence for either the sequential or the parallel EP.
When the likelihood terms are log-concave and the approximation is initialized to
the prior, the algorithm converges fine in many cases \citep[see
e.g.,][]{Nickisch+Rasmussen:2008}. However, with a non-log-concave likelihood such
as the Student-$t$ model, convergence problems may arise and these will be
discussed in section \ref{seq_EP_properties}. The convergence can be improved
either by damping the EP updates \citep{Minka:2002} or by using a robust but slower
double-loop algorithm \citep{Heskes:2002}. In damping the site parameters in their
natural exponential forms, $\tilde{\tau}_i = \tilde{\sigma}_i^{-2}$ and
$\tilde{\nu}_i = \tilde{\sigma}_i^{-2} \tilde{\mu}_i$, are updated to a convex
combination of the old and proposed new values, which results in the following
update rules:
\begin{equation} \label{EP_site_updates}
  \Delta \tilde{\tau}_i = \delta (\hat{\sigma}_i^{-2} -\sigma_i^{-2})
  \quad \text{and} \quad
  \Delta \tilde{\nu}_i = \delta (\hat{\sigma}_i^{-2} \hat{\mu}_i
  -\sigma_i^{-2} \mu_i ),
\end{equation}
where $\delta \in (0,1]$ is a step size parameter controlling the amount of
damping. Damping can be viewed as using a smaller step size within a gradient-based
search for saddle points of the same objective function as is used in the
double-loop algorithm \citep{Heskes:2002}.

\subsection{Expectation Propagation, the double-loop algorithm}

When either the sequential or the parallel EP does not converge one may still find
approximations satisfying the moment matching conditions (\ref{eq_moment_matching})
by a double loop algorithm. For example, \citet{Heskes:2002} present simulation
results with linear dynamical systems where the double loop algorithm is able to
find more accurate approximations when the damped EP fails to converge. For the
model under consideration, the fixed points of the EP algorithm correspond to the
stationary points of the following objective function \citep{Minka:2001b}
\begin{align}
   \min_{ \lamb_s } \max_{ \lamb_{-} } & \!
   - \sum_{i=1}^n \log \! \int \! p(y_i|f_i)
   \exp \left( \nu_{-i} f_i -\tau_{-i} \frac{ f_i^2}{2} \right) df_i
   - \log \! \int \! p(\f) \prod_{i=1}^n
   \exp \left( \tilde{\nu}_i f_i -\tilde{\tau}_i \frac{f_i^2}{2} \right) d\f
   \nonumber \\
   &+ \sum_{i=1}^n \log \int
   \exp\left( \nu_{s_i} f_i -\tau_{s_i} \frac{f_i^2}{2}  \right)  d f_i \label{EP_objective}
\end{align}
where $\lamb_{-}=\{\nu_{-i},\tau_{-i} \}$, $\tilde{\lamb}= \{ \tilde{\nu}_i,
\tilde{\tau}_i \}$, and $\lamb_s=\{\nu_{s_i},\tau_{s_i} \}$ are the natural
parameters of the cavity distributions $q_{-i}(f_i)$, the site approximations
$\tilde{t}_i(f_i)$, and approximate marginal distributions $q_{s_i}(f_i)=
\N(\tau_{s_i}^{-1} \nu_{s_i}, \tau_{s_i}^{-1})$ respectively. The min-max problem
needs to be solved subject to the constraints $\tilde{\nu}_i = \nu_{s_i} -
\nu_{-i}$ and $\tilde{\tau}_i = \tau_{s_i} - \tau_{-i}$, which resemble the moment
matching conditions in (\ref{eq_moment_matching}). The objective function in
(\ref{EP_objective}) is equal to the $\log Z_{\text{EP}}$ defined in
(\ref{eq_EP_post}) and is also equivalent to the expectation consistent (EC) free
energy approximation presented by \citet{Opper:2005}. A unifying view of the EC and
EP approximations as well as the connection to the Bethe free energies is presented
by \citet{Heskes:2005}.

Equation (\ref{EP_objective}) suggests a double-loop algorithm where the inner loop
consist of maximization with respect to $\lamb_-$ with fixed $\lamb_s$ and the
outer loop of minimization with respect to $\lamb_s$. The inner maximization
affects only the first two terms and ensures that the marginal moments of the
current posterior approximation $q(\f)$ are equal to the moments of the tilted
distributions $\hat{p}(f_i)$ for fixed $\lamb_s$. The outer minimization ensures
that the moments $q_{s_i}(f_i)$ are equal to marginal moments of $q(\f)$. At the
convergence, $q(f_i)$, $\hat{p}(f_i)$, and $q_{s_i}(f_i)$ share the same moments up
to the second order. If $p(y_i|f_i)$ are bounded, the objective is bounded from
below and consequently there exists stationary points satisfying these expectation
consistency constraints \citep{Minka:2001b,Opper:2005}. In the case of multiple
stationary points the solution with the smallest free energy can be chosen.
%
%

Since the first two terms in \eqref{EP_objective} are concave functions of
$\lamb_{-}$ and $\tilde{\lamb}$ the inner maximization problem is concave with
respect to $\lamb_-$ (or equivalently $\tilde{\lamb}$) after substitution of the
constraints $\tilde{\lamb} = \lamb_{s_i} - \lamb_{-} $ \citep{Opper:2005}. The
Hessian of the first term with respect to $\lamb_{-}$ is well defined (and negative
semi-definite) only if the tilted distributions $\hat{p}(f_i) \propto p(y_i|f_i)
q_{-i}(f_i)$ are proper probability distributions with finite moments up to the
fourth order. Therefore, to ensure that the product of $q_{-i}(f_i)$ and the
Student-$t$ site $p(y_i|f_i)$ has finite moments and that the inner-loop moment
matching remains meaningful,
%
%
the cavity precisions $\tau_{-i}$ have to be kept positive. Furthermore, since the
cavity distributions can be regarded as estimates for the leave-one-out (LOO)
distributions of the latent values, $\tau_{-i}=0$ would correspond to a situation
where $q(f_i|\y_{-i},\X)$ has infinite variance, which does not make sense given
the Gaussian prior assumption (\ref{GP_prior}). On the other hand, $\tilde{\tau}_i$
may become negative for example when the corresponding observation $y_i$ is an
outlier (see section \ref{seq_EP_properties}).

\subsection{Fractional EP updates}

Fractional EP \citep[or power EP,][]{Minka:2004} is an extension of EP which can be
used to reduce the computational complexity of the algorithm by simplifying the
tilted moment evaluations and to improve the robustness of the algorithm when the
approximation family is not flexible enough \citep{Minka:2005} or when the
propagation of information is difficult due to vague prior information
\citep{Seeger:2008}.
%
%
In the fractional EP the cavity distributions are defined as $q_{-i}(f_i) \propto
q(f_i|\mathcal{D}, \theta, \nu, \sigma^2)/\tilde{t}_i(f_i)^\eta$ and the tilted
distribution as $\hat{p}_i(f_i) \propto q_{-i}(f_i)p(y_i|f_i)^\eta$ for a fraction
parameter $\eta \in (0,1]$. The site parameters are updated so that the moments of
$q_{-i}(f_i)\tilde{t}_i(f_i)^\eta \propto q(f_i)$ match with $q_{-i}(f_i)
p(y_i|f_i)^\eta$. Otherwise the procedure is similar and the standard EP can be
recovered by setting $\eta=1$. In the fractional EP the natural parameters of the
cavity distribution are given by
\begin{equation} \label{fEP_cavity}
  \tau_{-i} =  \sigma_i^{-2} -\eta \tilde{\tau}_i
  \quad \text{and} \quad
  \nu_{-i} =  \sigma_i^{-2} \mu_i - \eta \tilde{\nu}_i,
\end{equation}
and the site updates (with damping factor $\delta$) by
\begin{equation} \label{fEP_update}
  \Delta \tilde{\tau}_i = \delta \eta^{-1} (\hat{\sigma}_i^{-2} -\sigma_i^{-2})
  \quad \text{and} \quad
  \Delta \tilde{\nu}_i = \delta \eta^{-1} (\hat{\sigma}_i^{-2} \hat{\mu}_i
  -\sigma_i^{-2} \mu_i ).
\end{equation}


The fractional update step $\min_q \KL( \hat{p}_i(f_i) || q(f_i))$ can be viewed as
minimization of another divergence measure called the $\alpha$-divergence with
$\alpha=\eta$ \citep{Minka:2005}.
%
%
Compared to the KL-divergence, minimizing the $\alpha$-divergence with $0<\alpha<1$
does not force $q(f_i)$ to cover as much of the probability mass of $\hat{p}(f_i)$
whenever $\hat{p}(f_i)>0$. As a consequence, the fractional EP tends to
underestimate the variance and normalization constant of
$q_{-i}(f_i)p(y_i|f_i)^\eta$, and also the approximate marginal likelihood
$Z_{EP}$. On the other hand, we also found that minimizing the KL-divergence in the
standard EP may overestimate the marginal likelihood with some data sets. In case
of multiple modes, the approximation tries to represent the overall uncertainty in
$\hat{p}_i(f_i)$ the more exactly the closer $\alpha$ is to 1. In the limit $\alpha
\rightarrow 0$ the reverse KL-divergence is obtained which is used in some form,
for example, in the fVB and KL approximations \citep{Nickisch+Rasmussen:2008}. Also
the double-loop objective function (\ref{EP_objective}) can be modified according
to the different divergence measure of the fractional EP
\citep{Cseke:2011,Seeger:2011}.

The fractional EP has some benefits over the standard EP with the non-log-concave
Student-$t$ sites. First, when evaluating the moments of
$q_{-i}(f_i)p(y_i|f_i)^\eta$, setting $\eta<1$ flattens the likelihood term which
alleviates the possible converge problems related to multimodality. This is related
to the approximating family being too inflexible and the benefits of different
divergence measures in these cases are considered by \citet{Minka:2005}. Second,
the fractional updates help to avoid the cavity precisions becoming too small, or
even negative. By choosing $\eta<1$, a fraction $(1-\eta) \tilde{\tau}_i$ of the
precision of the $i$:th site  is left in the cavity. This decreases the cavity
variances which in turn makes the tilted moment integrations numerically more
robust.
%
%
Problems related to cavity precision becoming too small can be present also with
log-concave sites when the prior information is vague. For example,
\citet{Seeger:2008} reports that with an underdetermined linear model combined with
a log-concave Laplace prior the cavity precisions remain positive but they may
become very small which induces numerical inaccuracies in the analytical moment
evaluations. These inaccuracies may accumulate and even cause convergence problems.
\citet{Seeger:2008} reports that fractional updates improve numerical robustness
and convergence in such cases.


\section{Robust implementation of the parallel EP algorithm}

The sequential EP updates are shown to be stable for models in which the exact site
terms (in our case the likelihood functions $p(y_i|f_i)$) are log-concave
\citep{Seeger:2008}. In this case, all site variances, if initialized to
non-negative values, remain non-negative during the updates. It follows that the
variances of the cavity distributions $q_{-i}(f_i)$ are positive and thus also the
subsequent moment evaluations of $q_{-i}(f_i)p(y_i|f_i)$ are numerically robust.
The non-log-concave Student-$t$ likelihood is problematic because both the
conditional posterior $p(\f|\D,\theta,\nu,\sigma)$ as well as the tilted
distributions $\hat{p}_i(f_i)$ may become multimodal.
Therefore extra care is needed in the implementation and these issues are discussed
in this section.



The double-loop algorithm is a rigorous approach that is guaranteed to converge to
a stationary point of the objective function (\ref{EP_objective}) when the site
terms $p(y_i|f_i)$ are bounded from below. The downside is that the double-loop
algorithm can be much slower than for example the parallel EP because it spends
much computational effort during the inner loop iterations, especially in the early
stages
%
%
when $q_{s_i}(f_i)$ are poor approximations for the true marginals. An obvious
improvement would be to start with damped parallel updates and to continue with the
double-loop method if necessary. Since in our experiments the parallel EP has
proven quite efficient with many easier data sets, we adopt this approach and
propose few modifications to improve the convergence in difficult cases. A parallel
EP initialization and a double-loop backup is also used by \citet{Seeger:2011} in
their fast EP algorithm.

The parallel EP can also be interpreted as a variant of the double-loop algorithm
where only one inner-loop optimization step is done by moment matching
(\ref{eq_moment_matching}) and each such update is followed by an outer-loop
refinement of the marginal approximations $q_{s_i}(f_i)$. The inner-loop step
consists of evaluating the tilted moments $\{ \hat{\mu}_i, \hat{\sigma}^2_i|
i=1,...,n\}$ with $q_{s_i}(f_i)=q(f_i)=\N(\mq_i,\Sq_{ii})$, updating the sites
(\ref{EP_site_updates}), and updating the posterior (\ref{eq_EP_post}). The
outer-loop step consists of setting $q_{s_i}(f_i)$ equal to the new marginal
distributions $q(f_i)$. Connections between the message passing updates and the
double-loop methods together with considerations of different search directions for
the inner-loop optimization can be found in the extended version of
\citep{Heskes:2002}. The robustness of the parallel EP can be improved by the
following modifications.
\begin{enumerate}
  \item After each moment matching step check that the objective
      ($\ref{EP_objective}$) increases. If the objective does not increase
      reduce the damping coefficient $\delta$ until increase is obtained. The
      downside is that this requires one additional evaluation of the tilted
      moments for every site per iteration, but if these one-dimensional
      integrals are implemented efficiently this is a reasonable price for
      stability.

  \item Before updating the sites (\ref{EP_site_updates}) check that the new
      cavity variances $\tau_{-i} = \tau_{s_i} - (\tilde{\tau}_i +\Delta
      \tilde{\tau}_i)$ are positive. If they are negative, choose a smaller
      damping factor $\delta$ so that $\tau_{-i}>0$. This computationally cheap
      precaution ensures that the increase of the objective
      (\ref{EP_objective}) can be verified according to the modification 1.
      %

  \item With modifications 1 and 2 the site parameters can still oscillate (see
      section \ref{seq_EP_properties} for an illustration) but according to our
      experiments the convergence is obtained with all hyperparameters values
      eventually. The oscillations can be reduced by updating $q_{s_i}(f_i)$
      only after the moments of $\hat{p}(f_i)$ and $q(f_i)$ are consistent for
      all $i=1,...,n$ with some small tolerance, for example $10^{-4}$.
      Actually, this modification results in a double-loop algorithm where the
      inner-loop optimization is done by moment matching
      (\ref{eq_moment_matching}). If no parallel initialization is done, often
      during the first 5-10 iterations when the step size $\delta$ is limited
      according to the modification 2, the consistency between $\hat{p}(f_i)$
      and $q(f_i)$ cannot be achieved. This is an indication of $q(\f)$ being
      too inflexible for the tilted distributions with the current
      $q_{s_i}(f_i)$. An outer-loop update $q_{s_i}(f_i)=q(f_i)$ usually helps
      in these cases.

  \item If no increase of the objective is achieved after an inner-loop update
      (modification 1), utilize the gradient information to obtain a better
      step size $\delta$. The gradients of (\ref{EP_objective}) with respect to
      the site parameters $\tilde{\nu}_i$ and $\tilde{\tau}_i$ can be
      calculated without additional evaluations of the objective function. With
      these gradients, it is possible to determine the gradient of the
      objective function with respect to $\delta$ in the current search
      direction defined by (\ref{EP_site_updates}). For example, cubic
      interpolation with derivative constraints at the end points can be used
      to approximate the objective as a function of $\delta$ with fixed site
      updates $\Delta \tilde{\tau}_i = \hat{\sigma}_i^{-2} -\sigma_i^{-2}$ and
      $\Delta \tilde{\nu}_i = \hat{\sigma}_i^{-2} \hat{\mu}_i -\sigma_i^{-2}
      \mu_i$ for $i=1,...,n$, from which a better estimate for the step size
      $\delta$ can be determined efficiently.


\end{enumerate}

In the comparisons of section \ref{seq_experiments} we start with 10 damped
($\delta=0.8$) parallel iterations because with a sensible hyperparameter
initialization this is enough to achieve convergence in most hyperparameter
optimization steps with the empirical data sets. If no convergence is achieved this
parallel initialization also speeds up the convergence of the subsequent
double-loop iterations (see section \ref{seq_convergence}). If after any of the
initial parallel updates the posterior covariance $\Sq$ becomes ill-conditioned,
i.e., many of the $\tilde{\tau}_i$ are too negative, or any of the cavity variances
become negative we reject the new site configuration and proceed with more robust
updates utilizing the previously described modifications. To reduce the
computational costs we limited the maximum number of inner loop iterations
(modification 3) to two with two possible additional step size adjustment
iterations (modification 4). This may not be enough to suppress all oscillations of
the site parameters but in practice more frequent outer loop refinements of
$q_{s_i}(f_i)$ were found to require fewer computationally expensive objective
evaluations for convergence.

In some rare cases, for example when the noise level $\sigma$ is very small, the
outer-loop update of $q_{s_i}(f_i)$ may result in negative values for some of the
cavity variances even though the inner-loop optimality is satisfied. In practise
this means that $[\Sq_{ii}]^{-1}$ is smaller than $\tilde{\tau}_i$ for some $i$.
This may be a numerical problem or an indication of a too inflexible approximating
family but switching to fractional updates helps. However, in our experiments, this
happened only when the noise level was set to too small values and with a sensible
initialization such problems did not emerge.


\subsection{Other implementation details}

The EP updates require evaluation of moments $m_k=\int f_i^k g_i(f_i) df_i$ for
$k=0,1,2$, where we have defined $g_i(f_i) =q_{-i}(f_i) p(y_i|f_i)^{\eta}$. With
the Student-$t$ likelihood and an arbitrary $\eta \in (0,1]$ numerical integration
is required. We used the adaptive Gauss-Kronrod quadrature described by
\citet{Shampine:2008} and for computational savings calculated the required moments
simultaneously using the same function evaluations. The integrand $\hat{p}_i(f_i)$
may have one or two modes between the cavity mean $\mu_{-i}$ and the observation
$y_i$. In the two-modal case the first mode is near $\mu_{-i}$ and the other near
$\mu_\infty=\sigma^2_{\infty} (\sigma_{-i}^{-2} \mu_{-i} + \eta_i \sigma^{-2}
y_i)$, where $\mu_{\infty}$ and $\sigma^2_{\infty} = (\sigma_{-i}^{-2} + \eta_i
\sigma^{-2})^{-1}$ correspond to the mean and variance of the limiting Gaussian
tilted distribution as $\nu \rightarrow \infty$. 
%
%
The integration limits were set to $\min(\mu_{-i} - 6\sigma_{-i},\mu_\infty -
6\sigma_{\infty})$ and $\max(\mu_{-i} + 6\sigma_{-i},\mu_\infty +
6\sigma_{\infty})$ to cover all the relevant mass around the both possible modes.




Both the hyperparameter estimation and monitoring the convergence of the EP
requires that the marginal likelihood $q(\y|\X,\theta,\sigma^2, \nu)$ can be
evaluated in a numerically robust manner. Assuming a fraction parameter $\eta$ the
marginal likelihood is given by
\begin{align}\label{eq_EP_marg_likelih}
  \log Z_{\text{EP}} =& \frac{1}{\eta} \sum_{i=1}^n
  \left( \log \hat{Z}_i + \frac{1}{2}
  \log \tau_{s_i} \tau_{-i}^{-1} + \frac{1}{2} \tau_{-i}^{-1} \nu_{-i}^2
  - \frac{1}{2} \tau_{s_i}^{-1} \nu_{s_i}^2 \right) \nonumber \\
  & -\frac{1}{2} \log|\I +\Kff \St^{-1}|
  -\frac{1}{2} \nut \Tr \mq,
\end{align}
where $\nu_{s_i}= \nu_{-i} +\eta \tilde{\nu}_{i}$ and $\tau_{s_i}= \tau_{-i} +\eta
\tilde{\tau}_{i}$. The first sum term can be evaluated safely if the cavity
precisions $\tau_{-i}$ and the tilted variances $\hat{\sigma}_i^2$ remain positive
during the EP updates because at convergence $\tau_{s_i}= \hat{\sigma}_i^{-2}$.

Evaluation of $|\I +\Kff \St^{-1}|$ and $\Sq=(\Kff ^{-1}+\St ^{-1})^{-1}$ needs
some care because many of the diagonal entries of $\St^{-1} =
\text{diag}[\tilde{\tau}_1,...,\tilde{\tau}_n]$ may become negative due to outliers
and thus the standard approach presented by \citet{Rasmussen+Williams:2006} is not
suitable. One option is to use the rank one Cholesky updates as described by
\citet{Vanhatalo+Jylanki+Vehtari:2009} or the LU decomposition as is done in the
GPML implementation of the Laplace approximation \citep{GPML}. In our parallel EP
implementation we process the positive and negative sites separately. We define
$\mb{W}_1=\text{diag}(\tilde{\tau}_i^{1/2})$ for $\tilde{\tau}_i>0$ and
$\mb{W}_2=\text{diag}( |\tilde{\tau}_i|^{1/2})$ for $\tilde{\tau}_i<0$, and divide
$\Kff$ into corresponding blocks $\mb{K}_{11}$, $\mb{K}_{22}$, and
$\mb{K}_{12}=\Tra{\mb{K}_{21}}$. We compute the Cholesky decompositions of two
symmetric matrices
\begin{align}\label{eq_L1}
  \mb{L}_1 \Tra{\mb{L}_1} = \I + \mb{W}_1 \mb{K}_{11} \mb{W}_1 \quad
  \text{and} \quad \mb{L}_2 \Tra{\mb{L}_2}
  = \I - \mb{W}_2 (\mb{K}_{22} -\mb{U}_2 \Tra{\mb{U}_2} ) \mb{W}_2, \nonumber
\end{align}
where $\mb{U}_2=\mb{K}_{21} \mb{W}_1 \mb{L}_1^{- \text{T}}$. The required
determinant is given by $|\I +\Kff \St^{-1}| = |\mb{L}_1|^2 |\mb{L}_2|^2$. The
dimension of $\mb{L}_1$ is typically much larger than that of $\mb{L}_2$ and it is
always positive definite. $\mb{L}_2$ may not be positive definite if the site
precisions are too negative, and therefore if the second Cholesky decomposition
fails after a parallel EP update we reject the proposed site parameters and reduce
the step size. The posterior covariance can be evaluated as $\Sq = \Kff - \mb{U}
\Tra{\mb{U}} + \mb{V} \Tra{\mb{V}} $, where $\mb{U} = [\mb{K}_{11}, \mb{K}_{12}]\Tr
\mb{W}_1 \mb{L}_1^{-\text{T}}$ and $\mb{V} = [\mb{K}_{12}, \mb{K}_{22}]\Tr \mb{W}_2
\mb{L}_2^{- \text{T}} - \mb{U} \Tra{\mb{U}_2} \mb{W}_2 \mb{L}_2^{- \text{T}}$. The
regular observations reduce the posterior uncertainty through $\mb{U}$ and the
outliers increase uncertainty through $\mb{V}$.

\section{Properties of the EP with a Student-t likelihood}
\label{seq_EP_properties}

In GP regression the outlier rejection property of the Student-$t$ likelihood
depends heavily on the data and the hyperparameters. If the hyperparameters and the
resulting unimodal approximation (\ref{eq_EP_post}) are suitable for the data there
are usually only a few outliers and there is enough information to handle them
given the smoothness assumptions of the GP prior and the regular observations. This
is usually the case during the MAP estimation if the hyperparameters are
initialized sensibly. Unsuitable hyperparameters values, for example a small $\nu$
combined with a too small $\sigma$ and a too large lengthscale, can result into a
very large number of outliers because the model is unable to explain large quantity
of the observations.
This may not necessarily induce convergence problems for the EP if there exists
only one plausible posterior hypothesis capable of handling the outliers. On the
other hand, if the conditional posterior distribution has multiple modes
convergence problems may occur unless sufficient amount of damping is used. In such
cases the approximating family \eqref{eq_EP_post} may not be flexible enough but
different divergence measures (fractional updates with $\eta<1$) can help
\citep{Minka:2005}. In this section we discuss the convergence properties of the EP
and also compare its approximation to the other methods described in the section 3.

An outlying observation $y_i$ increases the posterior uncertainty on the unknown
function at the input space regions a priori correlated with $\x_i$. The amount of
increase depends on how far the posterior mean estimate of the unknown function
value, $\E(f_i|\mathcal{D})$, is from the observation $y_i$. Some insight into this
behavior is obtained by considering the negative Hessian of $\log
p(y_i|f_i,\nu,\sigma^2)$, i.e., $W_i=-\nabla_{f_i}^2\log p(y_i|f_i)$, as a function
of $f_i$ (compare to the Laplace approximation in section 3.2). $W_i$ is
non-negative when $y_i - \sigma\sqrt{\nu} < f_i < y_i + \sigma \sqrt{\nu}$, attains
its minimum when $f_i = y_i \pm \sigma \sqrt{3\nu}$ and approaches zero as $|f_i|
\rightarrow \infty$. Thus, with the Laplace approximation $y_i$ satisfying
$\hat{f}_i - \sigma \sqrt{\nu} < y_i < \hat{f}_i +\sigma\sqrt{\nu}$ can be
interpreted as regular observations because they decrease the posterior covariance
$\mb{\Sigma}$ in equation (\ref{Hessian}). The rest of the observations increase
the posterior uncertainty and can therefore be interpreted as outliers.
Observations that are far from the mode $\hat{f}_i$ are clear outliers in the sense
that they have very little effect on the posterior uncertainty. Observations that
are close to $\hat{f}_i \pm \sigma \sqrt{3\nu}$ are not clearly outlying because
they increase the posterior uncertainty the most. The most problematic situations
arise when the hyperparameters are such that many $\hat{f}_i$ are close to $y_i \pm
\sigma \sqrt{3\nu}$. However, despite the negative $\mb{W}_{ii}$, the covariance
matrix $\Sq_{LA}$ is positive definite if $\f$ is a maximum of the conditional
posterior.

The EP behaves similarly as well. If there is a disagreement between the cavity
distribution $q_{-i}(f_i)= \N(\mu_{-i},\sigma_{-i}^2)$ and the likelihood
$p(y_i|f_i)$ but the observation is not a clear outlier, the tilted distribution is
two-modal and the moment matching (\ref{eq_moment_matching}) results in an increase
of the marginal posterior variance, $\hat{\sigma}_i^2>\sigma_i^2$, which causes
$\tilde{\tau}_i$ to decrease (\ref{EP_site_updates}) and possibly become negative.
%
%
The sequential EP usually runs smoothly when there's a unique posterior mode and
clear outliers. The site precisions corresponding to the outlying observations may
become negative but their absolute values remain small compared to the site
precisions of the regular observations.

%
%

\subsection{Simple regression examples}

\begin{figure}[t]
\centering
\includegraphics[width=0.95\columnwidth]{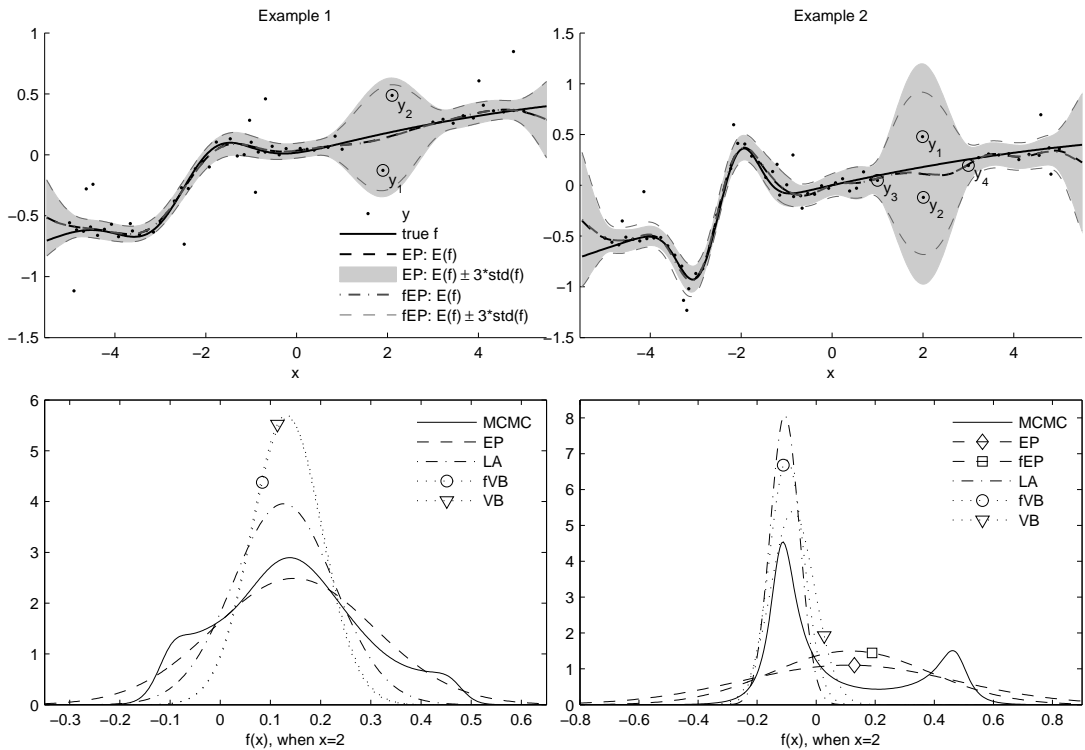}
\caption{The upper row: Two one-dimensional regression examples, where the standard EP
  may fail to converge with certain hyperparameter values, unless damped sufficiently.
  The EP approximations obtained by both the regular updates $\eta=1$ (EP) and the fractional
  updates $\eta=0.5$ (fEP) are visualized.
  The lower row: comparison of the approximative predictive distributions of the latent
  value $f(x)$ at $x=2$.
  With MCMC all the hyperparameters are sampled and for all the other approximations
  (except fVB in example 2, see the text for explanation)
  the hyperparameters are fixed to the corresponding MAP estimates.
  Notice that the MCMC estimate of the predictive distribution
  is unimodal in example 1 and multimodal in example 2. With
  smaller lengthscale values the conditional posterior $p(\f|\D,\theta)$
  can be multimodal also in example 1.}
\label{fig1}
\vskip -0.5cm
\end{figure}
%
%
%
Figure~\ref{fig1} shows two one-dimensional regression problems in which the
standard EP may run into problems. In example 1 (the upper left panel) there are
two outliers $y_1$ and $y_2$ providing conflicting information in a region with no
regular observations ($1<x<3$). If $\nu$ and $\sigma^2$ are sufficiently small, so
that the likelihood $p(y_i|f_i)$ is narrow as a function of $f_i$, and the
length-scale is small inducing small correlation between inputs far apart, there is
significant posterior uncertainty about the unknown $f(x)$ when $1<x<3$ and the
true posterior is multimodal. The Student-$t$ distribution is able to reject up to
$m$ outliers locally, in some neighborhood of an input $\x$, if there are at least
$2m$ observations in that neighborhood. The size of the neighborhood depends on the
smoothness properties of the GP prior (governed by the length-scale). In example 1,
we have two observations locally which both conflict with $q(f_1,f_2|y_3,...,y_n)$
almost equally much and neither one can be labeled as an outlier nor a regular
observation. Contrary to the clearly outlying observations for which
$\tilde{\tau}_i<0$, at convergence the site precisions $\tilde{\tau}_1$ and
$\tilde{\tau}_2$ (corresponding to $y_1$ and $y_2$) get small positive values, that
is, these observations decrease the posterior uncertainty despite being outliers.
If the site updates are not damped enough $\tilde{\tau}_1$ or $\tilde{\tau}_2$ may
become negative and cause stability problems.


If the length-scale is sufficiently large in the example 1, the GP prior forces the
posterior distribution to become unimodal and both the sequential and the parallel
EP converge. This is also the case when the hyperparameters are fixed to their MAP
estimates (assuming noninformative priors). The corresponding predictive
distribution is visualized in the upper left panel of Figure \ref{fig1} showing a
considerable increase in the posterior uncertainty when $1<x<3$. The lower left
panel shows comparison of the predictive distribution of $f(x)$ at $x=2$ obtained
with the different approximations described in the section 3. The hyperparameters
are estimated separately for each method. The smooth MCMC estimate is calculated by
integrating analytically over the latent vector $\f$ for each posterior draw of the
residual variances $\mathbf{V}$ and averaging the resulting Gaussian distributions
$q(\tilde{f}|\tilde{x},\mathbf{V},\theta)$. The MCMC estimate (with integration
over the hyperparameters) is unimodal but shows small side bumps when the latent
function value is close to the observations $y_1$ and $y_2$. The standard EP
estimate (EP1) covers well the posterior uncertainty on the latent value but both
the Laplace and fVB underestimate it. At the other input locations where the
uncertainty is small, all methods give very similar estimates.

The second one-dimensional regression example, visualized in the upper right panel
of Figure~\ref{fig1}, is otherwise similar to the example 1 except that the
nonlinearity of true function is much stronger when $-5<x<0$, and the observations
$y_1$ and $y_2$ are closer in the input space. The stronger nonlinearity requires a
much smaller length-scale for a good data fit and the outliers $y_1$ and $y_2$
provide more conflicting information (and stronger multimodality) due to the larger
prior covariance. The lower right panel shows comparison of the approximative
predictive distributions of $f(x)$ when $x=2$. The MCMC estimate has two separate
modes near the observations $y_1$ and $y_2$. The Laplace and fVB approximations are
sharply localized at the mode near $y_1$ but the standard EP approximation (EP1) is
very wide trying to preserve the uncertainty about the both modes. Contrary to the
example 1, also the conditional posterior $q(\f|\D,\theta,\nu,\sigma)$ is two-modal
if the hyperparameters are set to their MAP-estimates.

\begin{figure}[t]
  \begin{center}
    \subfigure[$q(f_1,f_2|y_3,...,y_n)$ and $p(y_1|f_1) p(y_2|f_2)$]{
      \includegraphics[width=0.22\columnwidth]{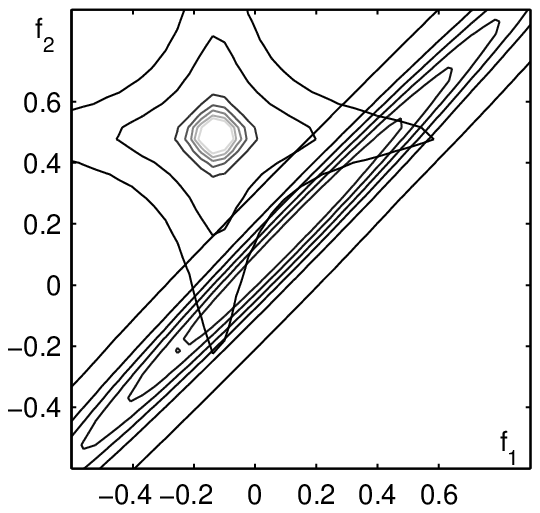}\label{fig_ep1}
    }
    \subfigure[$q(f_1,f_2|y_3,...,y_n)$ $\times p(y_1|f_1) p(y_2|f_2)$]{
      \includegraphics[width=0.22\columnwidth]{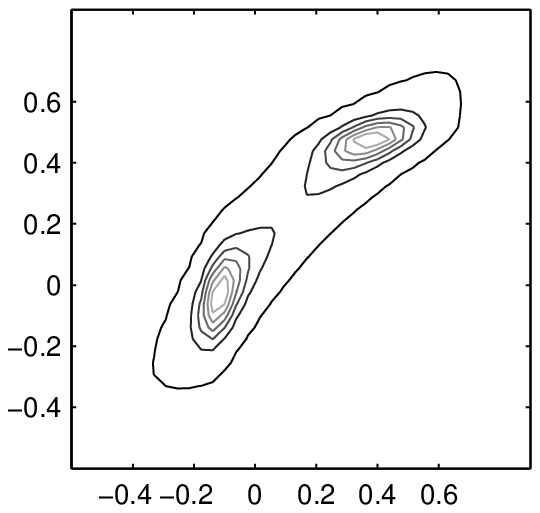}\label{fig_ep2}
    }
    \subfigure[$q(f_1,f_2|y_3,...,y_n)$ and $p(y_1|f_1)^{\eta} p(y_2|f_2)^{\eta}$]{
      \includegraphics[width=0.22\columnwidth]{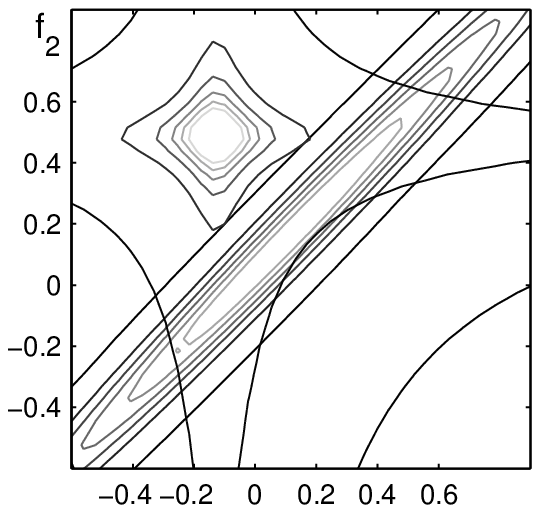}\label{fig_ep3}
    }
    \subfigure[$q(f_1,f_2|y_3,...,y_n)$ $\times p(y_1|f_1)^{\eta} p(y_2|f_2)^{\eta}$]{
      \includegraphics[width=0.22\columnwidth]{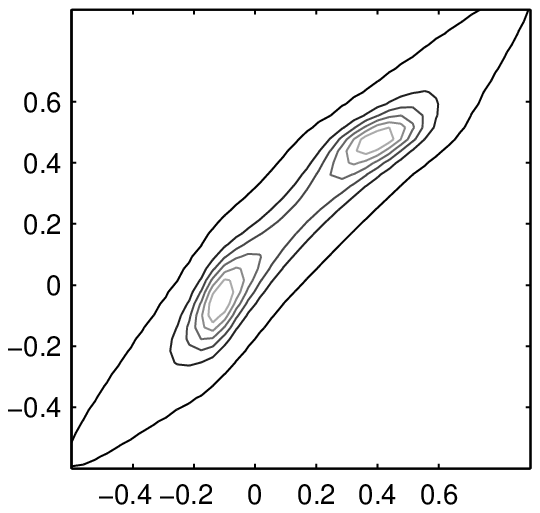}\label{fig_ep4}
    }
  \end{center}\caption{An illustration of a two-dimensional tilted
  distribution related to the two problematic data points $y_1$ and $y_2$ in the
  example 1. Compared to the MAP value used in the Figure \ref{fig1},
  shorter lengthscale (0.9) is selected so that the true conditional
  posterior is multimodal. Panel (a) visualizes the joint likelihood
  $p(y_1|f_1)p(y_2|f_2)$ together with the approximate marginal
  $q(f_1,f_2|y_3,...,y_n)$ obtained by one round of undamped sequential EP updates
  on sites $\tilde{t}_i(f_i)$, for $i=3,...,n$. Panel (b) visualizes the
  corresponding two-dimensional tilted distribution $\hat{p}(f_1,f_2) \propto
  q(f_1,f_2|y_3,...,y_n) p(y_1|f_1)p(y_2|f_2)$. Panels (c) and (d) show the same
  with only a fraction $\eta=0.5$ of the likelihood terms
  included in the tilted distribution, which corresponds to fractional EP updates
  on these sites.} \vskip -0.5cm \label{fig_ep}
\end{figure}

Next we discuss the problems with the standard EP updates with the help of the
example 1. Figure \ref{fig_ep} illustrates a two-dimensional tilted distribution of
the latent values $f_1$ and $f_2$ related to the observations $y_1$ and $y_2$ in
the example 1. A relatively small lengthscale (0.9) is chosen so that there is
large uncertainty on $f_1$ and $f_2$, but still quite strong prior correlation
between them. Suppose that all other sites have already been updated once with the
undamped sequential EP starting from a zero initialization ($\tilde{\tau}_i=0$ and
$\tilde{\nu}_i =0$ for $i=1,...,n$). Panel \subref{fig_ep1} visualizes the
2-dimensional marginal approximation $q(f_1,f_2|y_3,\ldots,y_n)$ together with the
joint likelihood $p(y_1,y_2|f_1,f_2) =p(y_2|f_2)p(y_2|f_2)$, and panel
\subref{fig_ep2} shows the contours of the resulting two dimensional tilted
distribution which has two separate modes. If the site $\tilde{t}_1(f_1)$ is
updated next in the sequential manner with no damping, $\tilde{\tau}_1$ will get a
large positive value and the approximation $q(f_1,f_2)$ fits tightly around the
mode near the observation $y_1$. After this, when the site $\tilde{t}_2(f_2)$ is
updated, it gets a large negative precision, $\tilde{\tau}_2<0$, since the
approximation needs to be expanded toward observation $y_2$ which is not at this
stage classified as a clear outlier because of the vague prior information. It
follows that during the second EP sweep site 1 can no longer be updated because the
cavity precision $\tau_{-1} =\sigma_1^{-2} -\tilde{\tau_{1}}$ is negative. This
happens because there are no other data points supporting the current posterior
solution, that is, there are no other sites with positive precision nearby, and the
site 2 with negative precision reduces the current marginal precision
$\sigma_1^{-2}$ too much. If the EP updates were done in parallel, both the cavity
and the site precisions would be positive after the first posterior update, but
$q(f_1,f_2)$ would be tightly centered between the modes. After a couple of
parallel loops over all sites, negative cavity variances can emerge as one of the
problematic sites gets a too small negative precision because the approximation
needs to be expanded to cover all the marginal uncertainty in the tilted
distributions.

Skipping updates on the sites with negative cavity variances can keep the algorithm
numerically stable (for an example of skipping updates see \citet{Minka:2002}), but
it is not enough to ensure convergence. In the Figure \ref{fig_ep}, the large
posterior uncertainty about $f_1$ and $f_2$ requires very small positive precisions
$\tilde{\tau}_i$ for sites 1 and 2. On the other hand, large differences between
the tilted and marginal variances induce large changes on these small site
precisions. If the EP updates are not constrained the site parameters may start
oscillating between too small and too large values and the algorithm never
converges because of too large update steps. One way to reduce the step length is
to use damped updates. Decreasing the damping factor $\delta$ sufficiently reduces
the oscillations so that the algorithm eventually converges but the convergence can
be very slow. Example 2 is more difficult in the sense that convergence requires
damping at least with $\delta=0.5$. With the sequential EP the convergence depends
also on the update order of the sites and $\delta<0.3$ is needed for convergence
with all permutations.

Figures \ref{fig_ep3}--\subref{fig_ep4} illustrate the same approximate tilted
distribution as Figures \ref{fig_ep1}--\subref{fig_ep2} but now only a fraction
$\eta=0.5$ of the likelihood terms are included. This corresponds to the first
round fractional updates on these sites with zero initialization. Because of the
flattened likelihood $p(y_1|f_1)^\eta p(y_2|f_2)^\eta$ the approximate posterior
distribution $q(f_i,f_2|y_3,...,y_n)$ is still two-modal but less sharply peaked
compared to the standard EP on the left. It follows that also the one-dimensional
tilted distributions have smaller variances and the consecutive fractional updates
(\ref{fEP_update}) of sites $1$ and $2$ do not widen the marginal variances
$\sigma_1^2$ and $\sigma_2^2$ as much. The site precisions $\tilde{\tau}_1$ and
$\tilde{\tau}_2$ are larger than with the regular EP and updates on them smaller
which requires less damping to keep them positive. This is possible because the
different divergence measure allows for a more localized approximation at $1<x<3$.
Figure \ref{fig1} shows a comparison of the standard (EP) and the fractional EP
(fEP, $\eta=0.5$) with MAP estimates of the hyperparameters. In the first example
both methods produce very similar predictive distribution because the posterior is
unimodal. In the second example (lower right panel) the fractional EP gives a much
smaller predictive uncertainty estimate when $x=2$ than the standard EP which in
turn puts more false posterior mass in the tails when compared to the MCMC.

\subsection{Convergence comparisons} \label{seq_convergence}

\begin{figure}[t]
\centering
\includegraphics[width=1\columnwidth]{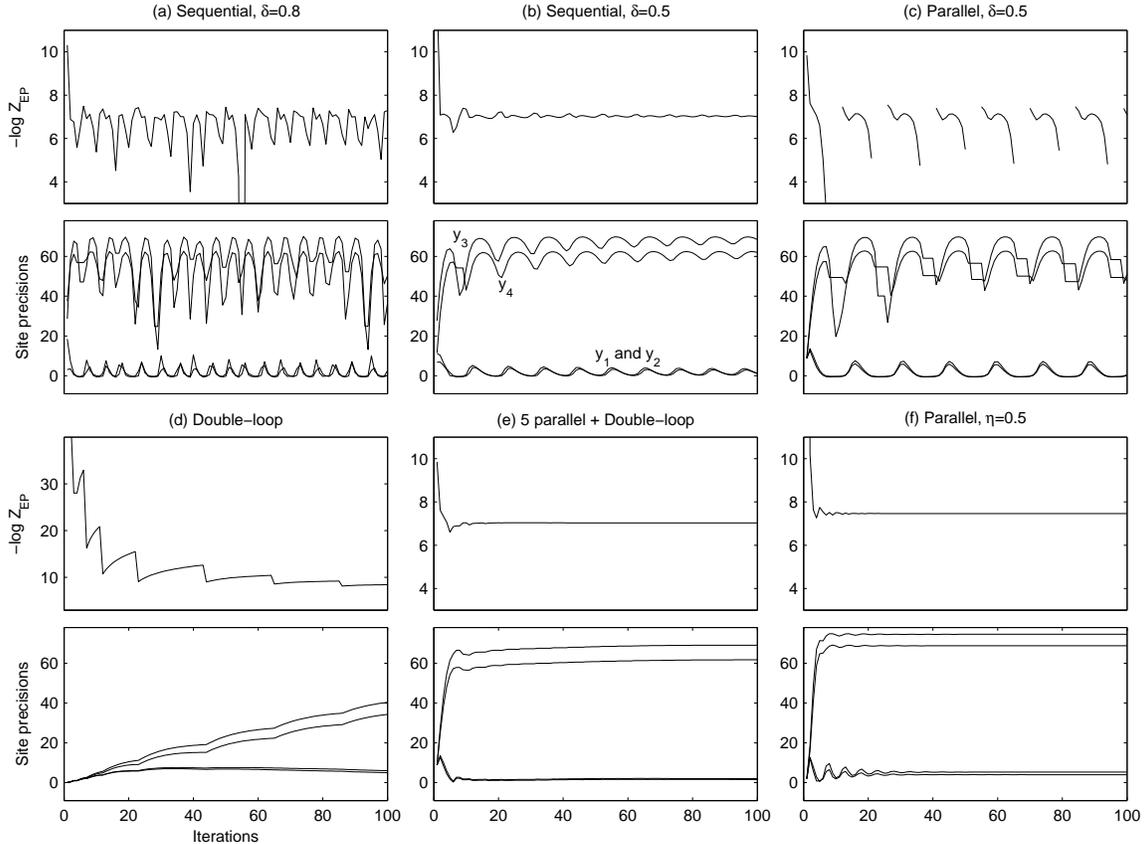}
\caption{A convergence comparison between the sequential and parallel EP as well as the
  double-loop algorithm in the example 2 (the right panel in Figure \ref{fig1}). For each
  method both the objective $-\log Z_{EP}$ and the site precisions $\tilde{\tau}_i$
  related to data points $y_1,...,y_4$ (see Figure \ref{fig1}) are shown.
  See the text for explanation.}
\label{fig_ep_convergence}
\end{figure}

Figure \ref{fig_ep_convergence} illustrates the convergence properties of the
different EP algorithms using the data from the example 2. The hyperparameters were
set to: $\nu=2$, $\sigma=0.1$, $\sigma_{se}=3$ and $l_k=0.88$. Panel (a) shows the
negative marginal likelihood approximation during the first 100 sweeps with the
sequential EP and the damping set to $\delta=0.8$. The panel below shows the site
precisions corresponding to the observations $y_{1},...,y_{4}$ marked in the upper
right panel of Figure \ref{fig1}. With this damping level the site parameters keep
oscillating with no convergence and there are also certain parameter values between
iterations 50-60 where the marginal likelihood is not defined because of negative
cavity precisions (the updates for such sites are skipped until next iteration).
Whenever $\tilde{\tau}_1$ and $\tilde{\tau}_2$ become too small they also inflict
large decrease in the nearby sites 3 and 4. These fluctuations affect other sites
the more the larger their prior correlations are (defined by the GP prior) with the
sites 1 and 2. Panel (b) shows the same graphs with larger amount of damping
$\delta=0.5$. Now the oscillations gradually decrease as more iterations are done
but convergence is still very slow. Panel (c) shows the corresponding data with the
parallel EP and same amount of damping. The algorithm does not converge and the
oscillations are much larger compared to the sequential EP. Also the marginal
likelihood is not defined at many iterations because of negative cavity precisions.

Panel (d) in Figure \ref{fig_ep_convergence} illustrates the convergence of the
double-loop algorithm with no parallel initialization. There are no oscillations
present because the increase of the objective (\ref{EP_objective}) is verified at
every iteration and sufficient inner-loop optimality is obtained before proceeding
with the outer-loop minimization. However, compared to the sequential or the
parallel EP, the convergence is very slow and it takes over 100 iterations to get
the site parameters to the level that the sequential EP attains with only a couple
of iterations. Panels (e) shows that much faster convergence can be obtained by
initializing with 5 parallel iterations and then switching to the double-loop
algorithm. There is still some slow drift visible in the site parameters after 20
iterations but changes in the marginal likelihood estimate are very small. Small
changes in the site parameters indicate differences in the moment matching
conditions (\ref{eq_moment_matching}) and consequently also the gradient of the
marginal likelihood estimate may be slightly inaccurate if the implicit derivatives
of $\log Z_{EP}$ with respect to $\lamb_-$ and $\lamb_s$ are assumed zero in the
gradient evaluations \citep{Opper:2005}. Panel (f) shows that the parallel EP
converges without damping if fractional updates with $\eta=0.5$ are applied.
Because of the different divergence measure the posterior approximation is more
localized (see Figure \ref{fig1}). It follows that the site precisions related to
$y_1$ and $y_2$ are larger and no damping is required to keep them positive during
the updates.


\subsection{The marginal likelihood approximation}

\begin{figure}[t]
\centering
\includegraphics[width=0.85\columnwidth]{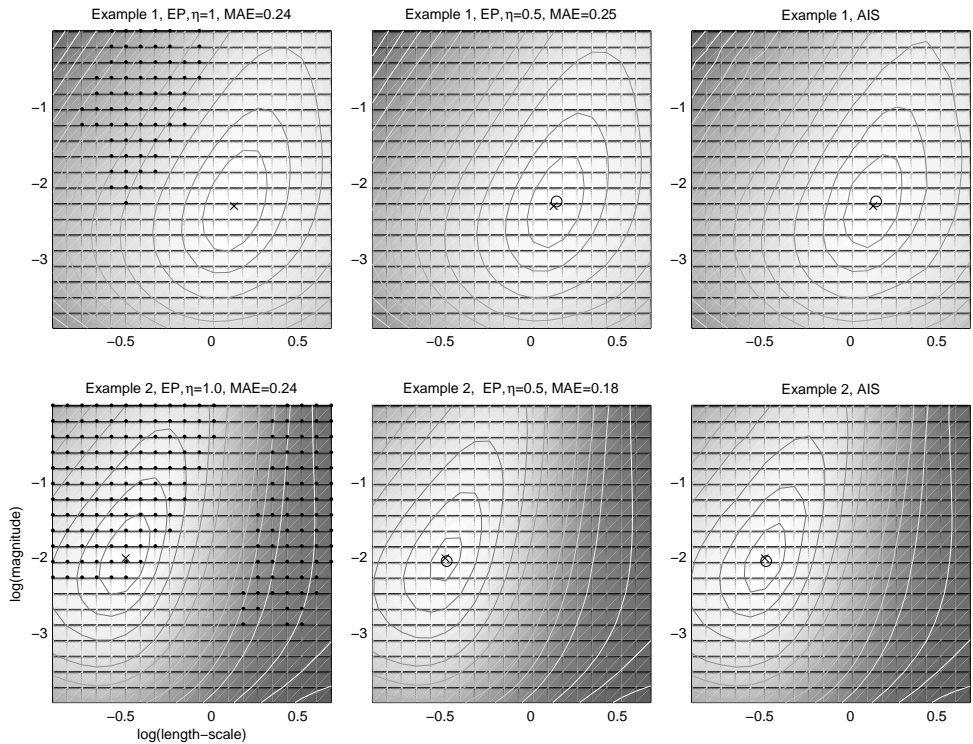}
\caption{The approximate log marginal likelihood $\log
  p(\y|\X,\theta,\nu,\sigma^2)$ as a function of the log-length-scale $\log(l_k^2)$
  and the log-magnitude $\log(\sigma_{se}^2)$ in the
  examples shown in the Figure \ref{fig1}. The marginal likelihood approximation is
  visualized with both the standard EP ($\eta=1$) and the fractional EP
  ($\eta=0.5$). The mode of the hyperparameters is marked with $\times$ and $\circ$
  for the standard and the fractional EP respectively. For comparison the marginal
  is also approximated by annealed importance sampling (AIS). For both the standard
  and the fractional EP the mean absolute errors (MAE) over the region
  with respect to the AIS
  estimate are also shown. The noise parameter $\sigma^2$ and the degrees of freedom
  $\nu$ are fixed to the MAP-estimates obtained with $\eta=1$. }
\label{fig3_marg}
\vspace{-0.5cm}
\end{figure}

Figure~\ref{fig3_marg} shows contours of the approximate log marginal likelihood
with respect to $\log(l_k)$ and $\log(\sigma^2_{se})$ in the examples of Figure
\ref{fig1}. The contours in the first column are obtained by applying first the
sequential EP with $\delta=0.8$ and using the double-loop algorithm if it does not
converge. The hyperparameter values for which the sequential algorithm does not
converge are marked with black dots and the maximum marginal likelihood estimate of
the hyperparameters is marked with ($\times$). The second column shows the
corresponding results obtained with the fractional EP ($\eta=0.5$) and the
corresponding hyperparameter estimates are marked with ($\circ$). For comparison,
log marginal likelihood estimates determined with the annealed importance sampling
(AIS) \citep{Neal:2001} are shown in the third column.


In the both examples there is an area of problematic EP updates with smaller
length-scales which corresponds to the previously discussed ambiguity about the
unknown function near data points $y_1$ and $y_2$ in the Figure \ref{fig1}. There
is also a second area of problematic updates at larger length-scale values in
example 2. With larger length-scales the model is too stiff and it is unable to
explain large proportion of the data points in the strongly nonlinear region
($-4<x<-1$) and consequently there exist no unique unimodal solution.
%
It is clear that with the first artificial example the optimization of the
hyperparameters with the sequential EP can fail if not initialized carefully or not
enough damping is used. In the second example the sequential EP approximation
corresponding to the MAP values cannot even be evaluated because the mode lies in
the area of nonconvergent hyperparameter values. In visual comparison with AIS both
the standard and fractional EP give very similar and accurate approximations in the
first example (the contours are drawn at the same levels for each method). In the
second example there are more visible differences: the standard EP tends to
overestimate the marginal likelihood due to the larger posterior uncertainties (see
Figure \ref{fig1}) whereas fractional EP underestimates it slightly. This is
congruent with the properties of the different divergence measure used in the
moment matching. The difference between the hyperparameter values at the modes
between the standard and fractional EP is otherwise less than 5\% except that in
the second example $\sigma$ and $\nu$ are ca. 30\% larger with the fractional EP.

\section{Experiments} \label{seq_experiments}

\begin{figure}[t]
\centering
\includegraphics[width=0.85\columnwidth]{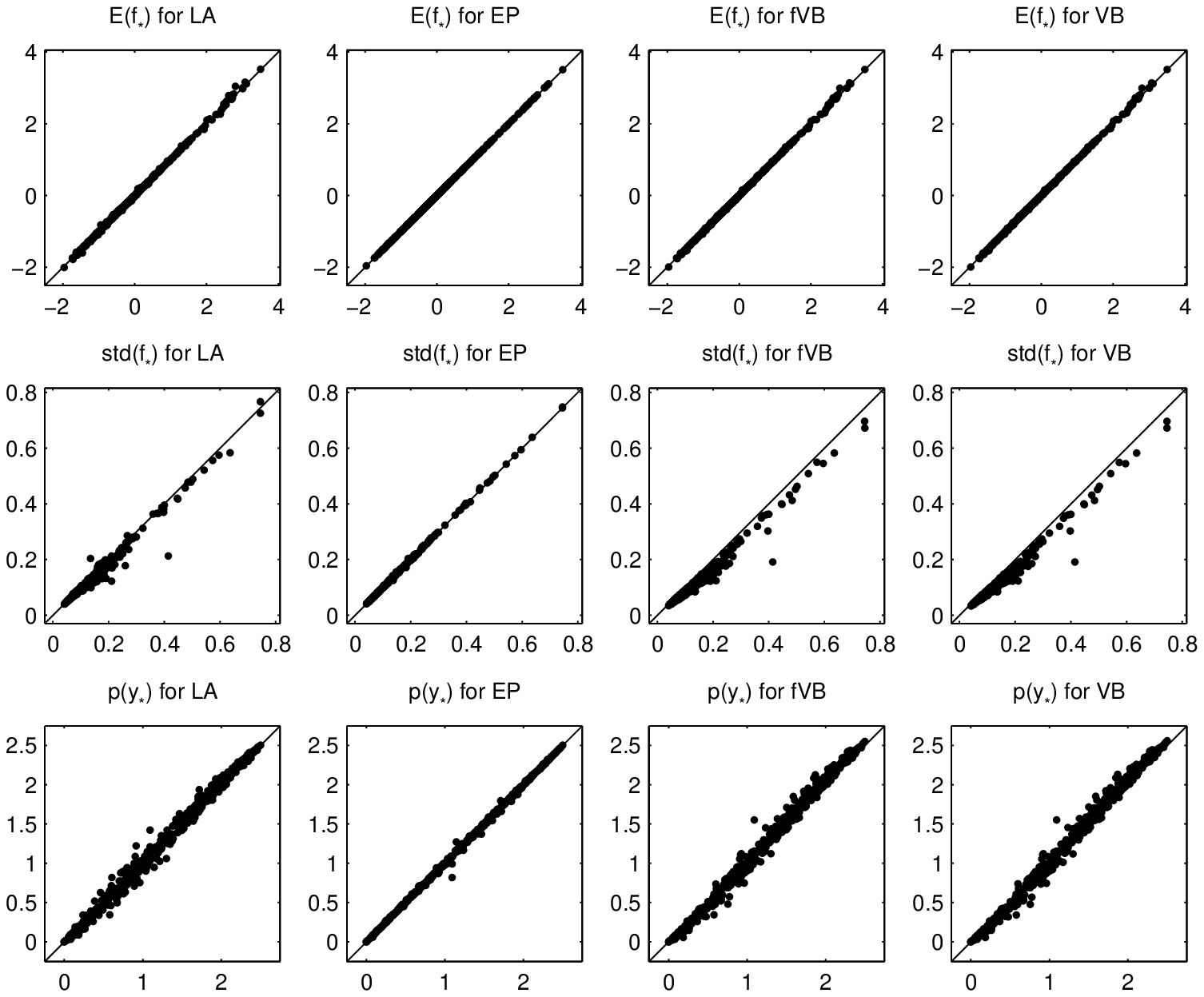}
\caption{A comparison of the approximative predictive means $\E(f_*|\x_*,\D)$,
standard deviations $\text{std}(f_*|\x_*,\D)$, and probabilities $q(y_*|\x_*,\D)$ provided
by the different approximation methods using 10-fold cross-validation on the Boston housing data.
The hyperparameters are fixed to the posterior means obtained by MCMC run on all data.
Each dot corresponds to one data point for which the x-coordinate is
the MCMC estimate with the fixed hyperparameter values and the y-coordinate the
corresponding approximative value obtained with the Laplace's method (LA), EP, fVB, or VB.}
\label{fig_comp_fixed}
\end{figure}

Four data sets are used to compare the approximative methods: 1) An artificial
example by \citet{Friedman:1991} involving a nonlinear function of 5 inputs. To
create a feature selection problem, five irrelevant input variables were added to
the data. We generated 10 data sets with 100 training points and 10 randomly
selected outliers as described by \citet{Kuss:2006}. 2) Boston housing data with
506 observations for which the task is to predict the median house prices in the
Boston metropolitan area with 13 input variables \citep[see e.g.,][]{Kuss:2006}. 3)
Data that involves prediction of concrete quality based on 27 input variables for
215 experiments \citep{Vehtari+Lampinen:2002}. 4) Data for which the task is to
predict the compressive strength of concrete based on 8 input variables for 1030
observations \citep{Cheng:1998}.



\subsection{Predictive comparisons with fixed hyperparameters}

First we compare the quality of the approximate predictive distributions
$q(\f_*|\x_*,\D,\theta,\nu,\sigma^2)$, where $x_*$ is the prediction location and
$f_* = f(x_*)$, between all the approximative methods. We run a full MCMC on the
housing data to determine the posterior mean estimates for the hyperparameters.
Then the hyperparameters were fixed to these values and a 10-fold cross-validation
was done with all the approximations including MCMC. The predictive means and
standard deviations of the latent values as well as the predictive densities of the
test observations obtained with the Laplace's method (LA), EP, fVB, and VB are
plotted against the MCMC estimate in the Figure \ref{fig_comp_fixed}. Excluding
MCMC, the predictive densities were approximated by numerically integrating over
the Gaussian approximation of $f_*$ in $q(y_*|\x_*,\D,\theta,\nu,\sigma^2) = \int
p(y_*|f_*,\nu,\sigma^2) q(f_*|\x_*,\D,\theta,\nu,\sigma^2) df_*$. EP gives the most
accurate estimates for all the predictive statistics, and clear differences to the
MCMC can only be seen in the predictive densities which indicates that accurate
mean and variance estimates of the latent value may not always be enough when
deriving other predictive statistics. This contrast somewhat to the corresponding
results in GP classification where Gaussian approximation was shown to be very
accurate in estimating predictive probabilities \citep{Nickisch+Rasmussen:2008}.
Both fVB and VB approximate the mean well but are overconfident in the sense that
they underestimate the standard deviations, overestimate the larger predictive
densities, and underestimate smaller predictive densities. LA gives similar mean
estimates with the VB approximations but approximates the standard deviations
slightly better especially with larger values. Put together, all methods provide
decent estimates with fixed hyperparameters but larger performance differences are
possible with other hyperparameter values (depending on the non-Gaussianity of the
true conditional posterior) and especially when the hyperparameters are optimized.


\subsection{Predictive comparisons with estimation of hyperparameter}

In this section we compare the predictive performance of LA, EP, fVB, VB, and MCMC
with estimation of the hyperparameters. The predictive performance was measured
with
the mean absolute error (MAE)
%
%
and the mean log predictive density (MLPD).
%
%
With the Friedman data these are evaluated using a test set of 1000 latent
variables for each of the 10 simulated data sets. For the Boston housing and
concrete quality data 10-fold cross validation is used. For the compressive
strength data 2-fold cross-validation was used because of the large number
observations. To assess the significance of the differences between model
performances, 95\% credible intervals for the MLPD measures were approximated by
Bayesian bootstrap as described by \citet{Vehtari+Lampinen:2002}. Gaussian
likelihood (GA) is selected as a baseline model for comparisons. With GA, LA, EP,
and VB the hyperparameters were estimated by optimizing the marginal posterior
densities whereas with MCMC all parameters were sampled. The fVB approach was
implemented following \citet{Kuss:2006} where the hyperparameters are adapted in
the M-step of the EM-algorithm. The variational lower bound associated with the
M-step was augmented with the same hyperpriors that were used with the other
methods.


Since the MAP inference on the degrees of freedom parameter $\nu$ proved
challenging due to possible identifiability issues, the LA, EP, fVB, and VB
approximations are tested with $\nu$ both fixed to $4$ (LA1, EP1, fVB1, VB1) and
optimized together with the other hyperparameters (LA2, EP2, fVB2, VB2). $\nu=4$
was chosen as a robust default alternative to the normal distribution which allows
for outliers but still has finite variance compared to the extremely wide-tailed
alternatives with $\nu \leq 2$. With EP we also tested a crude but very simple
approach (from now on EP3) for improving the robustness of the estimation of $\nu$.
We selected 15 values $\nu_j$ from the interval $[1.5, 20]$ linearly in the log-log
scale and ran the optimization of all the other hyperparameters with $\nu$ fixed to
these values. The conditional posterior of the latent values was approximated as
\begin{equation*}
  p(f_*|\D,\x) \approx \sum_j w_j q(\f| \D, \theta_j, \sigma_j^2, \nu_j),
\end{equation*}
where $\{ \theta_j, \sigma_j^2 \} = \argmax_{\theta,\sigma^2} q(\theta, \sigma^2 |
\D, \nu_j)$ and $w_j=q(\theta_j, \sigma_j^2, \nu_j|\D) / \sum_k q(\theta_k,
\sigma_k^2, \nu_k|\D)$.
%
%
This can be viewed as a crude approximation of the integration over $\nu$ where
$p(\theta,\sigma^2|\nu,\D)$ is assumed to be very narrowly distributed around the
mode. This approximation requires optimization of $\theta$ and $\sigma^2$ with all
the preselected values of $\nu$ and to speed up the computations $\theta$ and
$\sigma^2$ were initialized to the previous mode.

The squared exponential covariance (\ref{cf_sexp}) was used for all models. Uniform
priors were assumed for $\theta$ and $\sigma^2$ on log-scale and for $\nu$ on
log-log-scale.
%
%
The input and target variables were scaled to zero mean and unit variances. The
degrees of freedom $\nu$ was initialized to 4, $\sigma$ to 0.5 and the magnitude
$\sigma_{se}^2$ to 1. The optimization was done with different random
initializations for the length-scales $l_1,...,l_d$ and the result with the highest
posterior marginal density $q(\theta,\nu,\sigma^2|\mathcal{D})$ was chosen. The
MCMC inference on the latent values was done with both Gibbs sampling based on the
scale-mixture model (\ref{eq_scale_mixture}) and direct application of the scaled
HMC as described in \citet{Vanhatalo+Vehtari:2007}. Sampling of the hyperparameters
was tested with both slice sampling and HMC. The scale-mixture Gibbs sampling (SM)
combined with the slice sampling of the hyperparameters resulted in the best mixing
of the chains and gave the best predictive performance and therefore only those
results are reported. The convergence and quality of the MCMC runs was checked by
both visual inspections as well as by calculating the potential scale reduction
factors, effective number of independent samples, and autocorrelation times
\citep{Gelman+Carlin+Stern+Rubin:2004,Geyer:1992b}. Based on the convergence
diagnostics, burn-in periods were excluded from the beginning of the chains and the
remaining draws were thinned to form the final MCMC estimates.


%
\begin{figure}[!ht]
  \begin{center}
    \subfigure[Friedman]{
      \includegraphics[width=0.4\columnwidth]{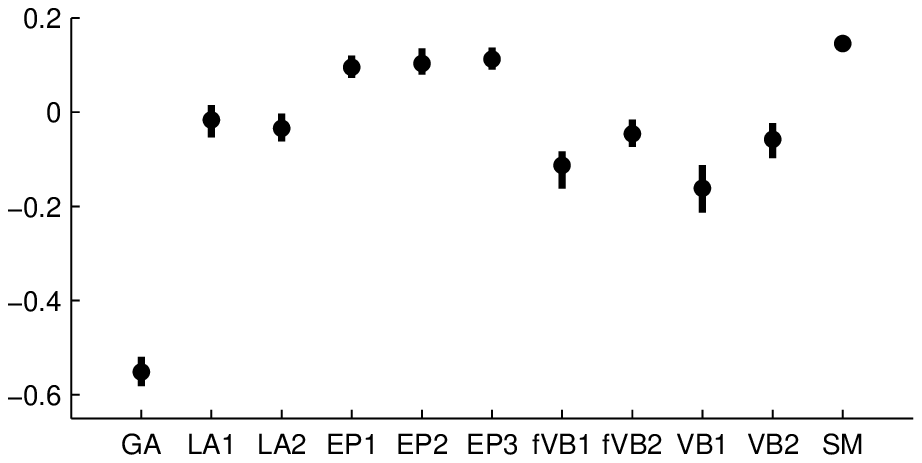}\label{friedman_lpd}
    }
    ~
    \subfigure[Friedman, pairwise]{
      \includegraphics[width=0.4\columnwidth]{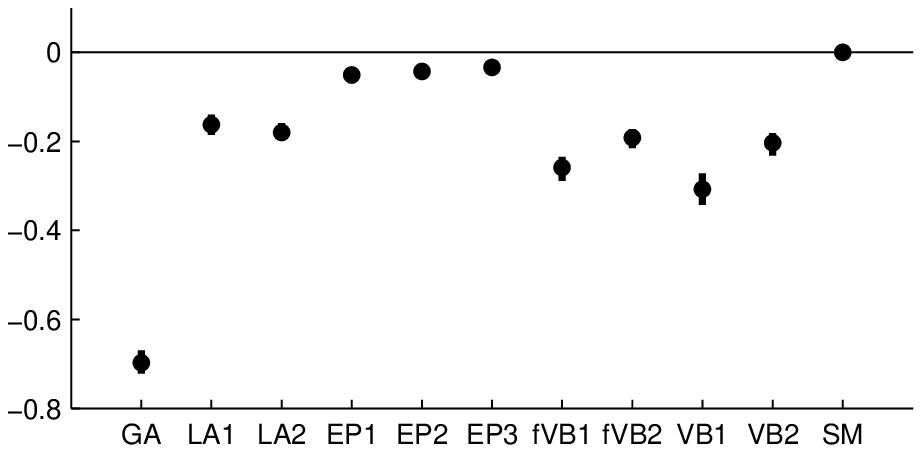}\label{friedman_complpd}
    }
    \subfigure[Boston housing]{
      \includegraphics[width=0.4\columnwidth]{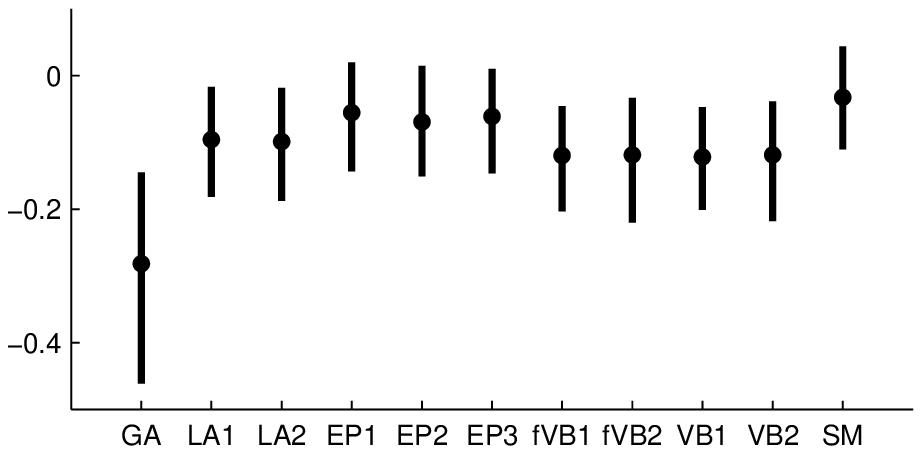}\label{housing_lpd}
    }
    ~
    \subfigure[Boston housing, pairwise]{
      \includegraphics[width=0.4\columnwidth]{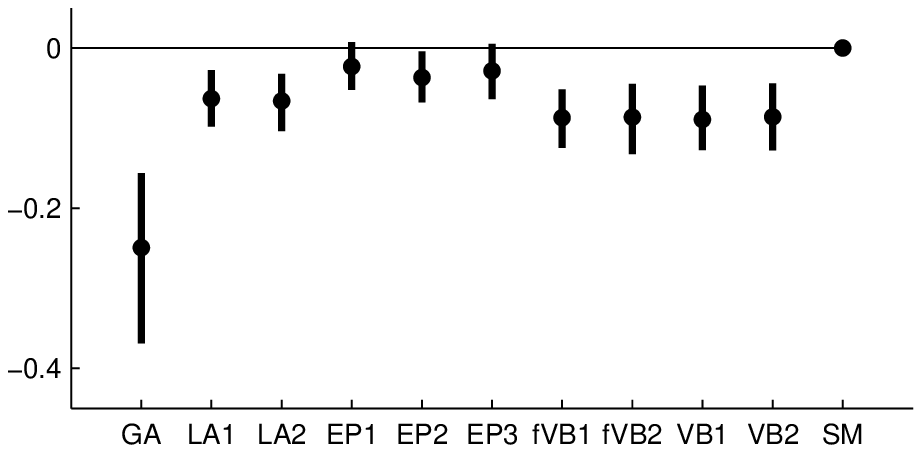}\label{housing_complpd}
    }
    ~
    \subfigure[Concrete quality]{
      \includegraphics[width=0.4\columnwidth]{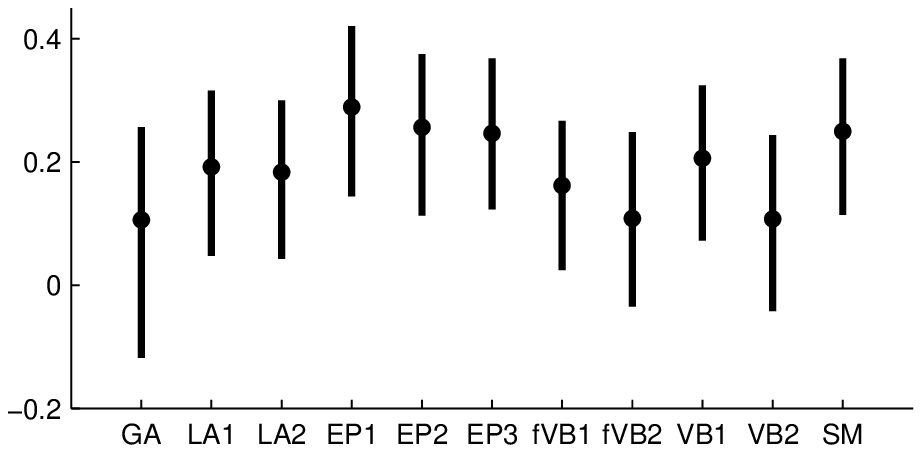}\label{concrete_lpd}
    }
    ~
    \subfigure[Concrete quality, pairwise]{
      \includegraphics[width=0.4\columnwidth]{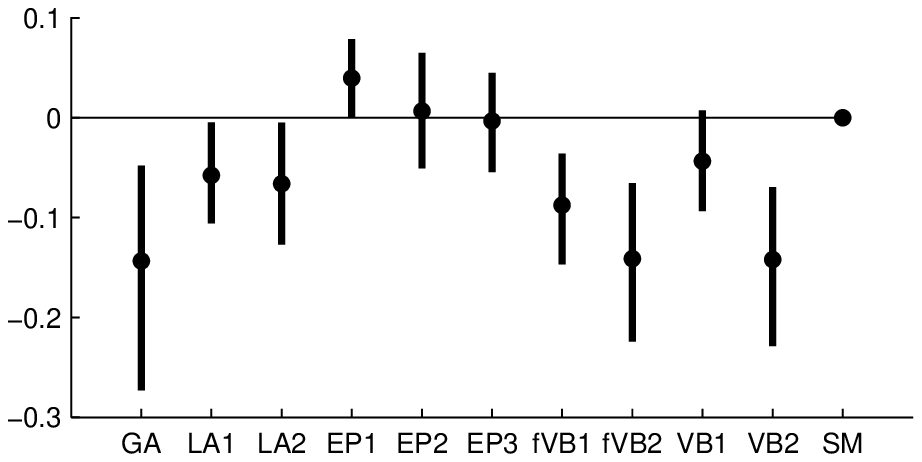}\label{concrete_complpd}
    }
    ~
    \subfigure[Compressive strength]{
      \includegraphics[width=0.4\columnwidth]{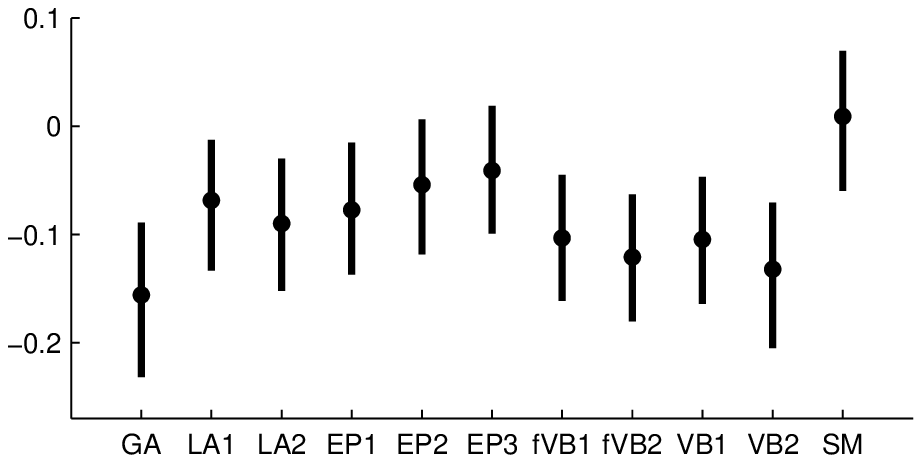}\label{concrete2_lpd}
    }
    ~
    \subfigure[Compressive strength, pairwise]{
      \includegraphics[width=0.4\columnwidth]{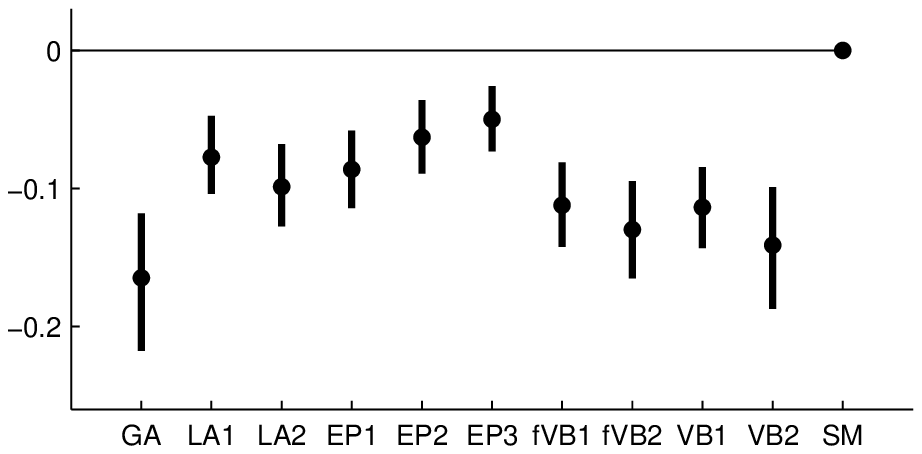}\label{concrete2_complpd}
    }
   \end{center}\caption{\textbf{Left column:} The mean log posterior
     predictive density (MLPD) and its 95\% central credible interval.
     Gaussian observation model (GA) is shown for reference. The Student-$t$ model
     is inferred with LA, EP, fVB, VB, and scale-mixture based Gibbs sampling (SM).
     Number 1 after a method means that $\nu$ is fixed, 2 means that it is optimized,
     and 3 stands for the simple approximative numerical integration over $\nu$.
     \textbf{Right column:}
     Pairwise comparisons of the log posterior predictive densities with respect to SM.
     The mean together with its 95\% central credible interval are shown.
     Values greater than zero indicate that a method is better than SM.}
  \label{LPD_results}
\end{figure}



Figures \ref{friedman_lpd}, \subref{housing_lpd}, \subref{concrete_lpd} and
\subref{concrete2_lpd} show the MLPD values and their 95\% credible intervals for
all methods in the four data sets. The Student-$t$ model performs better with all
approximations on all data sets compared to the Gaussian model. The differences
between the approximations are visible but not very clear. To illustrate the
differences more clearly Figures \ref{friedman_complpd}, \subref{housing_complpd},
\subref{concrete_complpd} and \subref{concrete2_complpd} show the pairwise
comparisons of the log posterior predictive densities to SM. The mean values of the
pairwise differences together with their 95\% credible intervals are visualized.
The Student-$t$ model with the SM implementation is significantly better than the
Gaussian model with higher than 95\% probability in all data sets. SM also performs
significantly better than all other approximations with the Friedman and
compressive strength data and with the Housing data only EP1 is not significantly
worse. The differences are much smaller with the concrete quality data and there
EP1 performs actually better than SM. One possible explanation for this is a wrong
assumption on the noise model (evidence for covariate dependent noise was found in
other experiments). Another possibility is the experimental design used in the data
collection; there is one input variable that is able to classify large proportion
of the observation with very small length scale, and sampling of this parameter may
lead to worse performance.

Other pairwise comparisons reveal that either EP1 or EP2 is significantly better
than LA, VB, and fVB in all data sets except the compressive strength data for
which significant difference is not found when compared to LA1. If the better
performing optimization strategy is selected from either LA, fVB, or VB, LA is
better than fVB and VB with the Friedman data and the compressive strength data.
Between fVB or VB no significant differences were found in pairwise comparisons.

Optimization of $\nu$ proved challenging and sensitive to the initialization of the
hyperparameters. The most difficult was fVB for which $\nu$ often drifted slowly
towards larger values.
%
This may be due to our implementation that was made following \citep{Kuss:2006} or
more likely to the EM style optimization of the hyperparameters. With LA, EP, and
VB the integration over $\f$ is redone in the inner-loop for all objective
evaluations in the hyperparameter optimization, whereas with fVB the optimization
is pursued with fixed approximation $q(\f|\D,\theta,\nu,\sigma^2)$. The EP-based
marginal likelihood estimates were most robust with regards to the hyperparameter
initialization. According to pairwise comparisons LA2 was significantly worse than
LA1 only in the compressive strength data. EP2 was significantly better than EP1 in
the housing and compressive strength data but significantly worse with the housing
data. With fVB and VB optimization of $\nu$ gave significantly better performance
only with the simulated Friedman data, and significant decrease was observed with
VB2 in the housing and compressive strength data. In pairwise comparisons, the
crude numerical integration over $\nu$ (EP3) was significantly better than EP1 and
EP2 with the housing and compressive strength data, but never significantly worse.
These results give evidence that the EP approximation is more reliable in the
hyperparameter inference because of the more accurate marginal likelihood estimates
which is in line with the results in GP classification
\citep{Nickisch+Rasmussen:2008}.




If MAE is considered the Student-$t$ model was significantly better than GA in all
other data sets except with the concrete quality data, in which case only EP1 gave
better results. If the best performing hyperparameter optimization scheme is
selected for each method, EP is significantly better than the others with all data
sets except with the compressive strength data in which case the differences were
not significant. When compared to SM, EP was better with the Friedman and concrete
quality data, otherwise no significant differences were found. LA was significantly
better than fVB and VB in the compressive strength data whereas with the simulated
Friedman data VB was better than LA and fVB.


\begin{table}
\centering
\caption{Two upper rows: The relative CPU times required for the hyperparameter
inference. The times are scaled to yield 1 for LA1 separately for each of the four
data sets, and both the relative mean (mean) as well as the maximum (max) over the
data sets are reported. The third row: The average relative CPU times over the four
data sets with the hyperparameters fixed to 28 preselected configurations.}
\begin{tabular}{l|c|c|c|c|c|c|c|c|c|c|c|}
      & GA & LA1 & LA2 & EP1 & EP2 & EP3 & fVB1 & fVB2 & VB1 & VB2 & SM \\
  \hline
  mean & 0.07 & 1.0 & 0.8 & 0.8 & 7.0 & 13 & 15 & 8.9 & 1.6 & 1.8 & 280 \\
  max & 0.09  & 1.0 & 1.2 & 1.1 & 16 & 26 & 39 & 22 & 3.3 & 3.8 & 440 \\
  \hline
  fixed & 0.1 & \multicolumn{2}{|c|}{1.0} & \multicolumn{3}{|c|}{5.5} &
    \multicolumn{2}{|c|}{2.4} & \multicolumn{2}{|c|}{1.9} & --
\end{tabular}
\label{tab_cpu}
\end{table}

Table \ref{tab_cpu} summarizes the total CPU times required for the posterior
inference (includes hyperparameter optimization and the predictions). The CPU times
are scaled to give one for LA1 and both the mean and maximum over the four data
sets are reported. The running times for the fastest Student-$t$ approximations are
roughly 10-fold compared to the baseline method GA. EP1, where $\nu=4$, is
surprisingly fast compared to the LA but with the optimization of $\nu$ (EP2) it
gets much slower. This is explained by the increasing number of double-loop
iterations required to achieve convergence with the larger number of difficult
posterior distributions as $\nu$ gets smaller values. The EP3 is clearly more
demanding compared to EP1 or EP2 because the optimization has to be repeated with
every preselected value of $\nu$. The fVB is quite slow compared to LA or VB
because of the slowly progressing EM-based hyperparameter adaptation. With the LA
and VB the running times are quite similar with $\nu$ both fixed and optimized. The
running times are suggestive in the sense that they depend much on the
implementations, convergence thresholds and initial guess of the hyperparameters.
Table \ref{tab_cpu} shows also the average relative running times over the four
data sets (excluding MCMC) with the hyperparameters fixed to 28 different
configurations (fixed). The configurations were created by first including the MCMC
mean for each data set and then generating all combinations of three clearly
different values of $\nu$, $\sigma$, and $\sigma_{se}$ around the MCMC mean with
randomly selected lengthscales. Quite many difficult hyperparameter configurations
were created which shows in the larger running time with the EP.


\section{Discussion}

Much research has been done on the EP and it has been found very accurate and
computationally efficient in many practical applications. Non-log-concave site
functions may be problematic for the EP but it has been utilized and found
effective for many potentially difficult models such as the Gaussian mixture
likelihoods \citep{Kuss:2006,Stegle+Fallert+MacKay+Brage:2008} and priors
\citep[spike and slab,][]{hernandez:2008}. Modifications such as the damping and
the fractional updates as well as alternative double-loop algorithms have been
proposed to improve the stability in difficult cases but the practical
implementation issues have not been discussed that much. In this work we have given
an example of the good predictive performance of the EP in a challenging model but
also analyzed the convergence problems and the proposed improvements from a
practical point of view. The Student-$t$ model is an interesting example because it
gives a natural interpretation for the negative site precisions as the increase of
the posterior uncertainty related to the not-so-clear outliers. On the other hand,
conflicting outliers in a region with considerable uncertainty about the unknown
function values may require very small but positive site precisions, which turned
out to be problematic for the regular EP updates in our experiments.

The nonlinear GP regression makes the inference problem even more challenging
because the multimodality of the conditional posterior depends on the
hyperparameter values. In practical applications, when the hyperparameters are
optimized, an estimate of the marginal likelihood is important also with the more
difficult cases. As we have demonstrated by examples, standard EP, unless damped
sufficiently, may not converge with the maximum marginal likelihood estimate of the
hyperparameters, and therefore, one cannot simply discard these hyperparameter
values. In our examples these situations were related to two modes in the
conditional posterior (caused by two outliers) quite far away from each other which
requires a very large local increase of marginal variance from the unimodal
posterior approximation. (It should also be noted that moderately damped sequential
EP worked fine with many other multimodal posterior distributions.) The globally
unimodal assumption is not the best in such cases although the true underlying
function is unimodal, but we think that it is important to get some kind of
posterior approximation. Whether one prefers the possible false certainty provided
by the Laplace or VB approximations, or the possible false uncertainty of the EP,
is a matter of taste but we prefer the latter one.

It is also important that the inference procedure gives some clue of the underlying
problems so that more elaborate models can be designed. In addition to the
examination of the posterior approximation, the need for double-loop iterations in
our EP implementation may be one indication of an unsuitable model. One can also
compare the cavity distributions which can be regarded as LOO estimates of the
latent values. If for certain sites most of the LOO information comes from the
corresponding site approximations, i.e., the cavity precisions are close to zero,
there is reason to suspect that the approximation is not suitable. Our EP
implementation enables a robust way of approaching this limit and in case of
problems one can switch to fractional updates.

In this work we have focused on the Student-$t$ model and have omitted comparisons
to other robust observations models such as the Laplace distribution and mixtures
of Gaussians \citep{Kuss:2006, Stegle+Fallert+MacKay+Brage:2008}. Laplace
distribution and Gaussian mixtures are computationally convenient since the moment
evaluations required for the EP updates can be done analytically. However, the
Student-$t$ model is more general in the sense that it can be thought of as an
infinite mixture of Gaussians and parameter $\nu$ can be used to continuously
adjust the degree of robustness. In addition, it contains the Gaussian distribution
as a special case. The thickness of the tails could be adjusted also with a finite
mixture of Gaussians but the number of parameters increases with the number of
mixture components. Inferring a point estimate of $\nu$ from the data turned out to
be challenging but EP was the most consistent approximative method for it in our
experiments. We showed that fixing $\nu=4$ gives a good baseline model for
comparisons and we also described a simple grid based approximation for improving
the estimation of $\nu$ based on data. The presented modifications for improving
the robustness of the parallel EP are general and they could be applied also for
other non-log-concave likelihoods. The presented EP approach for GP regression with
a Student-$t$ likelihood will be implemented in the freely available GPstuff
software package (\url{http://www.lce.hut.fi/research/mm/gpstuff/}).

\vskip 0.2in


\end{document}